\documentclass[preprint]{clv2025}

\jvol{vv}
\jnum{nn}
\jyear{2025}

\dochead{Short Paper} 

\usepackage{graphicx}
\usepackage{amsmath}
\usepackage{comment}
\usepackage{multirow}

\usepackage{cleveref}
\usepackage[dvipsnames]{xcolor}
\newcommand{\blue}[1]{{\color{blue}#1}}
\definecolor{darkblue}{rgb}{0, 0, 0.5}
\hypersetup{colorlinks=true,citecolor=darkblue, linkcolor=darkblue, urlcolor=darkblue}

\newcommand{\citeposs}[1]{\citeauthor{#1}'s \citeyearpar{#1}}

\runningtitle{Are Formal and Functional Linguistic Mechanisms Dissociated?}
\runningauthor{Hanna et al.}

\begin{document}

\title{Are Formal and Functional Linguistic Mechanisms Dissociated in Language Models?}
\author{Michael Hanna$^{1}$\thanks{Corresponding author}, Yonatan Belinkov$^{2}$, Sandro Pezzelle$^{1}$}

\affilblock{
    \affil{Institute for Logic, Language and Computation \\
  University of Amsterdam\\\quad \email{m.w.hanna@uva.nl}, \quad \email{s.pezzelle@uva.nl}}
    \affil{Technion -- Israel Institute of Technology\\\quad \email{belinkov@technion.ac.il}}
}
\maketitle

\begin{abstract}
Although large language models (LLMs) are increasingly capable, these capabilities are unevenly distributed: they excel at formal linguistic tasks, such as producing fluent, grammatical text, but struggle more with functional linguistic tasks like reasoning and consistent fact retrieval. Inspired by neuroscience, recent work suggests that to succeed on both formal and functional linguistic tasks, LLMs should use different mechanisms for each; such localization could either be built-in or emerge spontaneously through training. In this paper, we ask: do current models, with fast-improving functional linguistic abilities, exhibit distinct localization of formal and functional linguistic mechanisms? We answer this by finding and comparing the ``circuits'', or minimal computational subgraphs, responsible for various formal and functional tasks. Comparing 5 LLMs across 10 distinct tasks, we find that while there is indeed little overlap between circuits for formal and functional tasks, there is also little overlap between formal linguistic tasks, as exists in the human brain. Thus, a single formal linguistic network, unified and distinct from functional task circuits, remains elusive. However, in terms of cross-task faithfulness---the ability of one circuit to solve another's task---we observe a separation between formal and functional mechanisms, with formal task circuits achieving higher performance on other formal tasks. This suggests the existence of a set of formal linguistic mechanisms that is shared across formal tasks, even if not all mechanisms are strictly necessary for all formal tasks.
\end{abstract}

\section{Introduction}
A wide body of research has argued that language and thought are dissociated in the human brain \citep{Fedorenko2024TheLN}. That is, such research argues that the regions of the brain that respond differentially to well-formed linguistic input are distinct from those that respond to other structured inputs, such as mathematics, music, and code \citep{amalric2019distinct,chen2023music,ivanova2020comprehension}. They are also distinct from regions that respond to language-adjacent capabilities such as theory of mind and reasoning \citep{shain2022tom,monti2009boundaries}. These regions of the brain, termed the \textit{language network}, are thus selective for language (perhaps narrowly defined), and language alone.

In recent work, \citet{mahowald2024dissociating} argue that these two types of stimuli, to which the language network does and does not respond, correspond to two distinct types of linguistic competence: \textit{formal} and \textit{functional}. Formal linguistic competence is necessary for ``getting the form of language right''. It involves the correct structuring of language at the sub-word, lexical, and sentence level; in other words, phonology, morphology, syntax, and lexical semantics are all domains of formal linguistic competence. Past neuroscientific work has shown all of these to be supported by the language network \citep{regev2024high,shain2024distributed,hu2022precision}.

In contrast, functional linguistic competence involves abilities that allow speakers to achieve goals in the world, but may involve non-linguistic cognition. For example, speakers may engage in formal or pragmatic reasoning, or employ world knowledge in conversation, without such abilities being intrinsic to language. Similarly, speakers may use situation modeling skills to make sense of a narrative, or engage in theory of mind reasoning to understand their interlocutors' point of view, although language use need not entail the use of these abilities. Moreover, exercising these abilities does not engage the brain's language network.

\citeauthor{mahowald2024dissociating} furthermore claim that this distinction between formal and functional linguistic competence is reflected in the performance of today's large language models (LLMs). In particular, LLMs have strong formal linguistic competence as evidenced by their strong performance on syntax benchmarks and their general ability to output fluent and natural text \citep{hu-etal-2020-systematic,gauthier-etal-2020-syntaxgym,warstadt-etal-2020-blimp-benchmark}. However, their functional linguistic competence is markedly worse: LLMs frequently output false reasoning, hallucinate untrue facts, and fail at complex social reasoning \citep{dziri2023faith,xu2024hallucination,strachan2024testing}.

Much work has attempted to ameliorate these problems via retrieval-augmented generation or the use of chain-of-thought reasoning \citep{gao2024rag,wei2022cot}. However, \citeauthor{mahowald2024dissociating} offer another solution: perhaps LLMs would have stronger functional linguistic abilities if formal and functional linguistic abilities were as distinct in LLMs as they are in the human brain. Such a dissociation in LLMs could come about in two ways: either it could be explicitly built into them, or it could arise naturally via training, or the model's inductive biases. Today's LLMs have no explicit formal-functional modularity; however, it remains unknown whether formal-functional modularity has nonetheless arisen.

In this paper, we seek to answer that question: to what extent are formal and functional linguistic competence dissociated in the internals of today's LLMs?  If the two are not dissociated, new architectures or training procedures that bias models towards a formal-functional dissociation may be necessary to achieve this. If the two are already dissociated (despite the fact that LLMs struggle more with functional linguistic competence), this might indicate that dissociation does not suffice to improve LLMs' abilities. This question of dissociation is also relevant due to the increase in work that uses LLMs to explicitly model language in the human brain, oftentimes by predicting activations within the brain using those from LLMs \citep{tuckute2024language,sucholutsky2024getting}. The presence or absence of a dissociated language network in LLMs could help us judge whether such comparisons are licensed. Rather than comparing activations within models and brains, though, we propose to characterize mechanisms within models, and verify whether they are organized in the same way as the human brain.

To investigate whether formal and functional competence are distinctly localized within the LLMs, we draw on techniques from LLM interpretability, and in particular, \textit{mechanistic interpretability}, which studies the mechanisms that underlie LLMs' behavior using low-level causal methods \citep{ferrando2024primer}. Concretely, we study the formal-functional distinction using \textbf{circuits}: a circuit is a small path through the LLM that contains all of the mechanisms underlying its behavior on a task of interest \citep{olah2020zoom,wang2023interpretability}. Circuits provide causal guarantees that the localization found is correct, as all parts of the model outside of the circuit can be ablated without changing model behavior; see \Cref{sec:circuits} for more details. This targeted, causal evidence for the correctness of our localization is a notable advantage of our framework, as such evidence is harder to come by in human brains.\footnote{In humans, one must rely on either natural experiments (e.g., individuals with brain damage) or less precise causal methodologies. These are numerous, ranging from more targeted methods such as transcranial magentic stimulation or focused ultrasound, to the more general interventions used in psychedelics research, but they are overall less precise than our causal interventions; optogenetics allows for finer-grained causal manipulation of neurons, but only in non-human subjects.}

\begin{figure*}
    \centering
    \includegraphics[width=\textwidth]{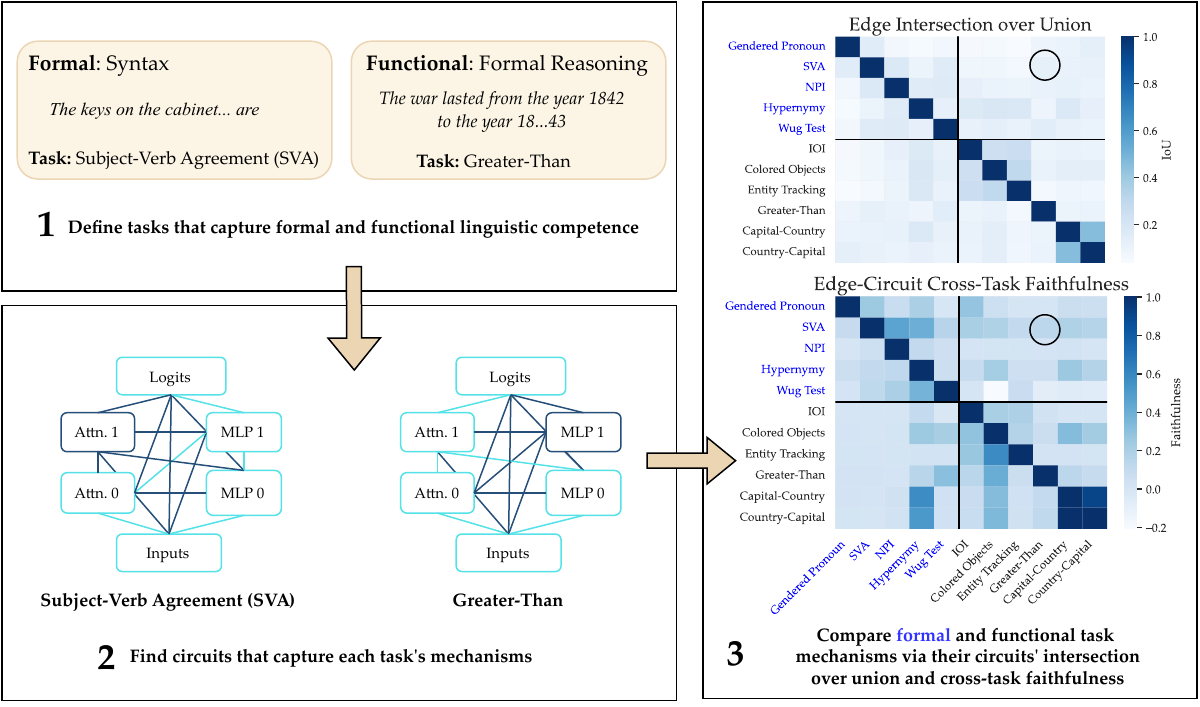}
    \caption{Our experimental pipeline. First, we define tasks that capture various aspects of formal or functional linguistic linguistic competence. Then, we find circuits for these tasks, which describe the model mechanisms responsible for them. Finally, we compare those circuits, measuring the similarity between \blue{formal} and functional task mechanisms (as measured by e.g. their circuits' intersection over union or cross-task faithfulness).}
    \label{fig:first-figure}
\end{figure*}

We thus translate \citeauthor{mahowald2024dissociating}'s hypothesis about emergent dissociation between formal and functional linguistic abilities in LLMs into the LLM circuit analysis framework. First, we identified 5 tasks involving formal linguistic competence, and 5 tasks involving functional linguistic competence. We next selected 5 LLMs, and found the circuits responsible for their behavior in these tasks. We then measured the similarity between each pair of task circuits, focusing in particular on similarities both within and across the formal and functional task groups; see \Cref{fig:first-figure} for an overview. 

But what does it mean for the two task circuits to be similar or dissociated? One way to operationalize this is to measure the overlap between formal and functional circuits, in terms of the components (and connections between them). If formal and functional networks are dissociated, we expect that formal and functional circuits should have low overlap, while formal circuits should have high overlap with one another. Our findings suggest formal and functional language competence are not dissociated in LLMs when this is measured via circuit overlap. Circuits for formal tasks have small but non-zero overlap with circuits for functional tasks. More importantly, circuits for formal tasks do not have an especially high overlap with one another. 

We also measure similarity dissociation via the ability of one circuit to perform another's task, or \textit{cross-task faithfulness}, as past work has found measuring overlap and cross-task faithfulness to yield different results \citep{hanna2024have}. In this setting, one circuit is similar to another if it can perform the other's task well. Thus, if formal and functional competence are dissociated in LLMs, we would expect formal task circuits to perform formal tasks well, but perform functional tasks poorly (and vice-versa). Under this metric, a formal-functional divide appears more plausible: formal task circuits indeed perform other formal tasks better than they perform functional ones; moreover, functional task circuits perform other functional tasks better than they perform formal ones. We conclude that although a unified formal network in terms of overlap does not exist, formal language circuits are indeed more similar to each other in terms of the tasks that they can perform.

In summary, we find evidence for emerging formal-functional dissociation in LLMs, though only in terms of cross-task faithfulness. That is, formal task circuits perform other formal tasks better than they perform functional ones; functional tasks circuits similarly struggle to perform formal tasks. However, our overlap studies suggest against the existence of one unified region necessary for all language tasks. Rather, there may be a broad pool of mechanisms shared across formal tasks; some formal tasks may be solved by multiple mechanisms (circuits) from this pool. In performing these analyses, we apply mechanistic interpretability techniques to a question from neuroscience in LLMs for the first time. We moreover conduct a study of circuits across LLMs, employing both the greatest number and largest size LLMs of such a study to date. We release the code for our experiments, as well as the efficient circuit-finding tools within them, at \url{https://github.com/hannamw/formal-functional-dissociation}.

\section{Background}
\subsection{The Language Network in the Human Brain}\label{sec:language-network}
The language network, as described by \citet{Fedorenko2024TheLN}, has three main characteristics: (1) it responds in an undifferentiated fashion to various types of language use; (2) it is consistent across modalities and languages; (3) and it is robustly dissociated from both low- and high-level networks with non-language roles. 

That is, the language network responds differentially to language as opposed to non-language stimuli, but it responds equally to syntax as it does to (lexical) semantics; the two do not activate distinct regions of the brain \citep{fedorenko2020selectivity,shain2024distributed}. Moreover, the language network activates on both linguistic input and output \citep{menenti2011shared}. It activates whether the language is heard or read, and is similar across languages \citep{regev2013selective,malik2022investigation}. However, it does not overlap with perceptual or motor regions, and also excludes regions for higher-level non-linguistic competence, such as cognitive control and theory of mind. \citep{li2024demistifying,pritchett2018highlevel,quillen2021distinct}.

The bulk of these findings come from experiments using a relatively simple setup used to localize brain regions responsible for a given competence. In such studies, brain activations (typically measured using functional magnetic resonance imaging; (fMRI) are measured in response to two contrasting stimuli (or while performing two contrasting tasks), where only one is relevant to the phenomenon of interest. Areas that differentially respond to the phenomenon of interest are inferred to be involved in its processing. Using these methods, \citet{Fedorenko2010NewMF} localize the language network by finding regions that activate on well-formed sentences as opposed to lists of non-words. Note that the \textit{differential} response requirement ensures that low-level regions that are activated by both contrasting stimuli are excluded from the localized region: for example, the language network excludes low-level speech processing areas, as these are activated by both sentences and non-word lists.

In summary, there is a wide body of evidence for the formal-function dissociation. Though this dissociation is not universally supported \citep{murphy2024LN,forkel2024redefining}, the evidence in its favor is substantial, spanning over a decade of research, and consisting of studies using a wide variety of languages and non-linguistic tasks. We believe this body of neuroscientific evidence provides a solid theoretical motivation for our study on whether emergent formal-functional modularity, as suggested by \citeauthor{mahowald2024dissociating}, exists in today's LLMs. However, see \Cref{sec:discussion} for discussion of dissenting views, and how our results might be interpreted from alternative perspectives.

\subsection{Formal and Functional Linguistic Competence in Language Models}
Building on the literature regarding the language network in the brain, \citet{mahowald2024dissociating} propose a related distinction between \textit{formal} and \textit{functional} linguistic competence. They use \textit{formal linguistic competence} to refer to linguistic abilities necessary to get the \textit{form} of language right; \textit{formal linguistic competence} refers to language abilities that we use to achieve goals or otherwise \textit{function} with language. Notably, while formal linguistic competence cleanly maps onto the language network, functional linguistic competence includes only some non-language-network abilities; others, like music, are excluded. 

\begin{table*}[]
    \includegraphics[width=\textwidth]{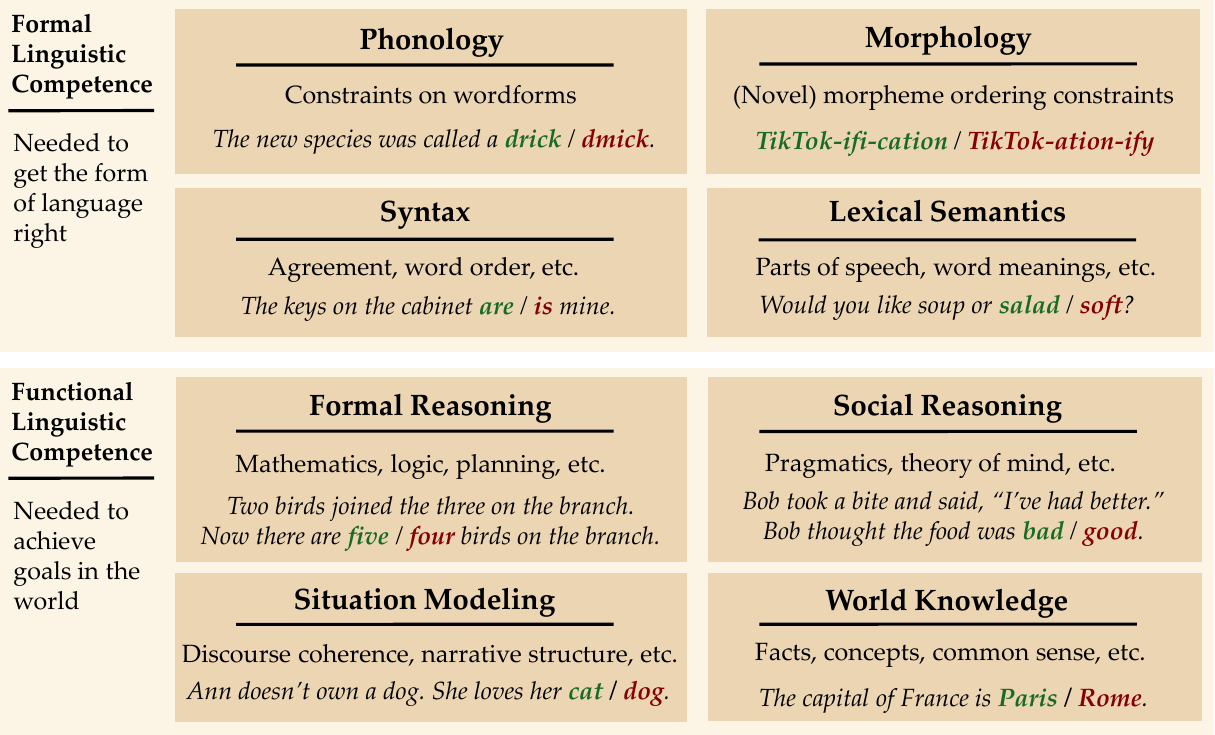}
    \caption{Subdomains of formal and functional linguistic competence. Each subdomain includes a sentence that is  {\color{OliveGreen} \textbf{correct}} or {\color{BrickRed}\textbf{incorrect}} with respect to it, depending on the word chosen. Figure adapted from \citet{mahowald2024dissociating}}
    \label{tab:mahowald-table}
\end{table*}

\citeauthor{mahowald2024dissociating} divide formal and functional linguistic competence into subdomains (\Cref{tab:mahowald-table}). Formal language competence includes subdomains like \textbf{phonology}, \textbf{morphology}, \textbf{syntax}, and \textbf{lexical semantics}. For \citeauthor{mahowald2024dissociating}, phonology corresponds to the rules governing valid wordforms (i.e., phonotactics), while morphology involves the correct ordering of morphemes. Syntax involves not only correct word order, but also higher-level abilities like agreement (e.g. between subjects and verbs). Lexical semantics entails using words correctly according to their part of speech, lexical category, or meaning. \citeauthor{mahowald2024dissociating} distinguish this category from semantics more broadly: general conceptual knowledge belongs to functional language competence.

In contrast, functional language competence consists of \textbf{formal reasoning}, \textbf{world knowledge}, \textbf{situation modeling}, and \textbf{social reasoning}. Formal reasoning includes math and logical abilities, while world knowledge includes facts and commonsense knowledge. Situation modeling entails the ability to track the state of a discourse, and the structure of narratives. Finally, social reasoning covers pragmatics and theory of mind.

\citeauthor{mahowald2024dissociating} note that LLMs have strong formal linguistic competence, succeeding on tests of syntactic ability and lexical semantics \citep{chang-bergen-2024-language}. It is more challenging to measure English LLMs' abilities in phonology, given that LLMs seldom produce novel phonemes, and morphology, given the relative simplicity of English morphology, but LLMs do seem to generate valid novel morphemes \citep{mccoy-etal-2023-much}. In contrast, models often struggle with functional tasks \citep{dziri2023faith,strachan2024testing}, though recent LLMs have markedly improved reasoning abilities due to intensive post-training. Still, there remains a clear gap between the relative ease of learning formal linguistic competence, and the ongoing challenge of functional linguistic competence.

The solution, according to \citeauthor{mahowald2024dissociating}, is to induce modularity in LLMs, just as it exists in the human brain. Such modularity could take two forms: \textit{architectural modularity}, which is explicitly built into a model's architecture, and \textit{emergent modularity}, which occurs naturally due to e.g. the model's inductive bias and training process. They note, as do we, that transformers are well-suited for this sort of emergent modularity: past work has found LLM attention heads (imperfectly) dedicated to certain syntactic relations \citep{vig-belinkov-2019-analyzing,clark-etal-2019-bert}, as well as heads that add as induction, succession, and copy suppression modules \citep{olsson2022context,gould2024successor,mcdougall-etal-2024-copy}. How, though, can we localize such modules if they exist?

\subsection{Causal Localization in LLMs}\label{sec:causal-localization}
Localizing the regions of a model that perform a given task or ability is a key question in the interpretability of NLP models. Modern interpretability work often uses causal interventions \citep{pearl2009causality} to do so. The core idea behind these is that a unit---e.g., a parameter in a weight matrix, or neuron in a model activation---is important if perturbing (or \textit{intervening on}) it causes the relevant model behavior to change. For example, if setting a neuron's activation to zero in a model causes the model to be unable to recall a country's capital, we conclude that the neuron played a role in the model's capital-recall ability. Causal interventions thus allow us to infer the function of units within a model.

A wide body of work performs localization in the parameter space of models. As zeroing-out parameters one by one is prohibitively expensive, it is common to learn a binary mask over parameters, indicating which parameters are important \citep{han2015learning,frankle2018lottery,prasanna-etal-2020-bert}. Unimportant parameters are set to zero; such masks are learned to maximize both sparsity and model performance under this regime. These techniques were initially developed to increase model efficiency, but have since been used to locate modules within models \citep{csordas2021are}, find language-specific and knowledge-critical subnetworks \citep{lin-etal-2021-learning,choenni-etal-2023-cross,bayazit-etal-2024-discovering}.

Other work instead localizes mechanisms in activation space, looking for neurons, components (such as transformer models' attention heads or multi-layer perceptrons), or entire layers that are important to task abilities. Such work has measured whole layers' importance by perturbing them with Gaussian noise \citep{meng2022locating}, zeroed out entire attention head activations \citep{voita-etal-2019-analyzing, olsson2022context}, or computed linear approximations of ablation effects \citep{nanda2023attribution}.

However, one must be cautious when performing causal interventions. For one, performing the right intervention is essential. While zero ablations are common and intuitive, they are harmful because model activations (and parameters) are seldom zero; zeroing them out may bring the activations out of distribution, causing harm \citep{hase2021out,chan2022causal}. Thus, if we zero ablate a given unit and observe a drop in model performance, we cannot determine if this stems from the unit's importance, or the out-of-distribution issue \citep{li2024optimal}. Mean ablations, which replace activations with their mean across a dataset, are less harmful, but can still fall out-of-distribution. Activation patching, which intervenes on a model by replacing a component's activation on one example, with an activation on another example, avoids this issue \citep{vig2020causal,geiger2021causal}; to our knowledge, no similar technique has been developed for parameter localization.

Moreover, causal interventions of this sort tell us only which units are \textit{necessary}; they  do not prove that the localized units are \textit{sufficient} to perform the task of interest, or that we have captured all relevant units. This issue and the issue with zero ablations have cast some doubt on older causal localization studies. Fortunately, recent work has developed a framework for the causal localization of mechanisms in transformer models that avoids many of these problems: circuits.

\section{Circuits}\label{sec:circuits}
We characterize the mechanisms behind LLMs' formal and functional linguistic competence using circuits \citep{olah2020zoom,elhage2021mathematical}. Circuits are small subgraphs of a model---generally no more than 5\% thereof---that capture how it performs a given task. Crucially, circuits are both necessary and sufficient for models to perform tasks; they destroy task performance when ablated, and suffice to perform the task when everything outside them is ablated. That is, circuits aim to capture entire task mechanisms. We propose to compare the similarity of task mechanisms by comparing their circuits.

\subsection{Definitions} 
A model's \textbf{circuit} for a given task is the minimal computational subgraph of the model that is faithful to its behavior on the task \citep{wang2023interpretability,hanna2023how}. That is, even if all parts of the model outside of the circuit are ablated (or instead corrupted), model behavior will not change. See \citet{hanna2024have} or \citet{miller2024transformercircuitfaithfulnessmetrics} for reviews of the circuits literature.

\paragraph{Computational Graph}
In this study, we focus on LLMs using the transformer architecture \citep{vaswani2017attention}. Such a model's computational graph describes the computations it performs, and can be viewed as a directed graph that begins at the model's inputs, flows through its intermediate components, and ends at its logits \citep{conmy2023towards, hanna2023how}. In this paper, the intermediate components we study are either individual attention heads or multi-layer perceptrons (MLPs); thus, every node in our graph is an input (token) node, attention head, MLP, or an output (logit) node.

\begin{figure*}
    \centering
    \includegraphics[width=\textwidth]{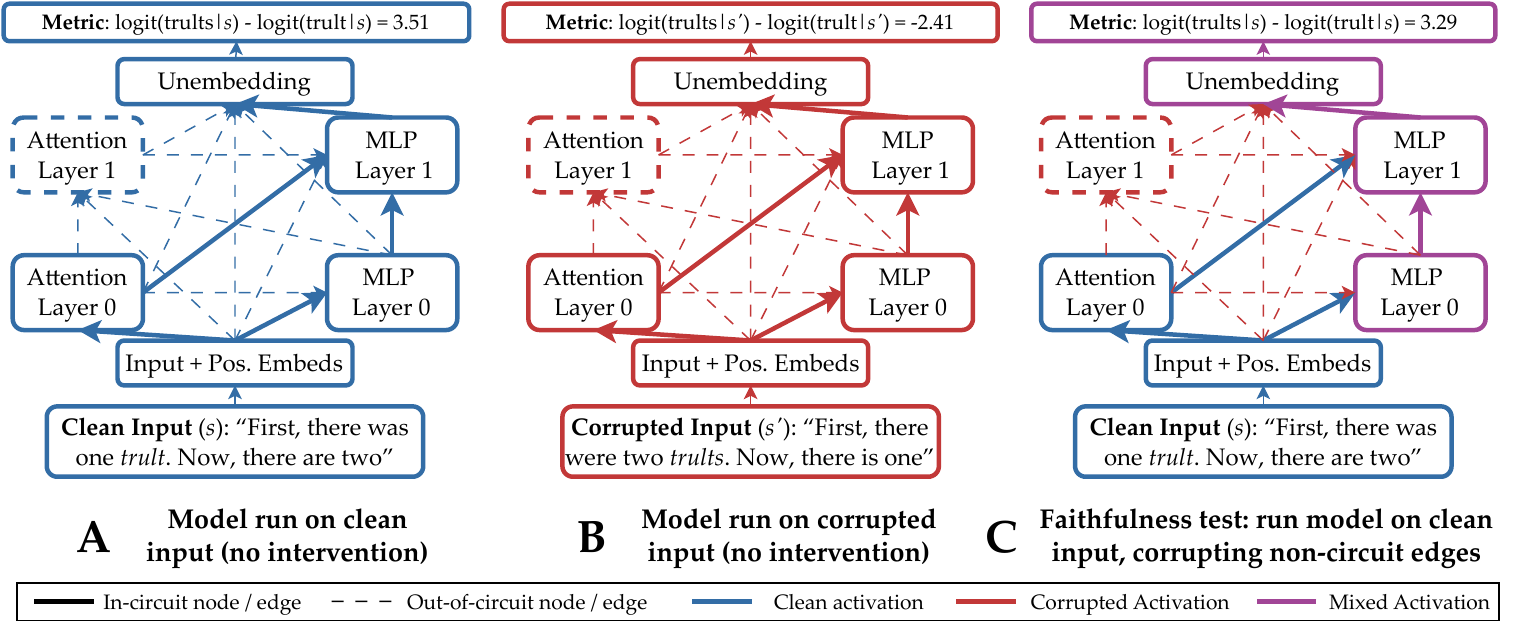}
    \caption{A toy circuit in a 2-layer transformer, where nodes are attention and MLP layers. \textbf{A}: We first run our whole model on the clean inputs to establish baseline behavior. \textbf{B}: We then run the model on a corrupted input that elicits very different behavior, and save the corrupted activations. \textbf{C}: To test a circuit's faithfulness, we run the model on clean inputs, replacing non-circuit node / edge activations with corrupted activations; behavior should stay the same as in (A).}
    \label{fig:toy-circuit}
\end{figure*}

Most autoregressive transformers can be conceptualized as having a \textit{residual stream}. This residual stream is initialized with the input embeddings (and potentially positional embeddings); then, each component takes the stream as input, and adds its own output to this stream. As a result, each component's input is the sum of the outputs of previous components. This means that each component has a direct effect (unmediated by other components) to every component downstream of it; it also has an indirect effect (mediated by other components). Edges in our computational graph represent these direct effects, so every component has an edge to all components that come later in the models. The input to a given node $v$ is the sum of the output of all nodes $u$ with an edge to $v$. A circuit should identify those edges that are important for the model's ability to perform the task at hand. See \Cref{fig:toy-circuit} for a toy example of a circuit in a 2-layer transformer model's computational graph.

\paragraph{Task} A \textit{task} in circuits analysis consists of clean and corrupted inputs, expected outputs, and a metric to measure task performance. For example, a subject-verb agreement task (like that of \citealp{linzen-etal-2016-assessing,newman-etal-2021-refining}) might consist of clean inputs like $s=$ \textit{The keys on the cabinet}; the expected output would be a plural-conjugated verb, like \textit{are}. Each clean input is paired with a corrupted input, which is drawn from the same task distribution, but is crafted to elicit very different behavior from the model. Here, the corrupted input might be $s'=$\textit{The key on the cabinet}, which elicits a singular-conjugated verb like \textit{is}. $s$ and $s'$ should have the same length in tokens.

Our metric could then be the difference in probabilities assigned to correctly and incorrectly conjugated verbs, given $s$; i.e., $m = p(\text{plural-verb}|s) - p(\text{singular-verb}|s)$. Note that $m$ should be high when the model is run on $s$, but low (negative) when it is run on $s'$, i.e. $s'$ elicits the opposite model behavior compared to $s$ with respect to $m$.

The metric should be continuous, both for use with circuit-finding methods (see \Cref{sec:circuit-finding}) and to allow for observation of incremental changes in model behavior as changes are made to the circuit. Tasks should be solvable by the model in which the circuit is to be found; if the model cannot solve the task, there may not be any task behavior to localize. The model's performance on a given task example should ideally be measurable in one forward pass.

\paragraph{Faithfulness} A circuit is \textit{faithful} if the model's task behavior stays the same, even when all nodes and edges outside the circuit are \textit{corrupted}. At a high level, corruption entails running our model on clean inputs $s$, but replacing the activations of the nodes and edges outside the circuit, with activations taken from corrupted inputs $s'$. As the vast majority of the model is outside the circuit, the model will behave as if it were being run on $s'$, and the value of the metric will drop---unless we have correctly localized all task-relevant nodes and edges.

More formally, and following \citet{hanna2024have}, we test model faithfulness by running the model on $s$ and performing the following causal intervention. Let $v$ be a node in the whole model's computational graph $G=(V,E)$. Let $z_{u}$ denote the activation of a given node $u$ during the current forward pass, and $z'_{u}$ denote its activation on corrupted inputs $s'$. Let $C=(V_C,E_C)$ denote our circuit. 

For each non-input node $v$, we set its input to
\begin{equation} 
v_{in} = \sum_{(u,v)\in E_C}  z_u + \sum_{(u,v)\in E\setminus E_C} z'_u.
\end{equation}
If all edges into $v$ are in the circuit, its input is $\sum_{(u,v)\in E} z_u$, the same as its input without interventions; if none are, its input is the same as it is when the model is run on $s'$.

Having done this, we measure $m_C$, the circuit's performance. Denote by $m$ the model's original performance on the task, and by $m_\emptyset$ the model's performance when entirely corrupted (i.e., when it is run on the corrupted input). We can then compute a measure of normalized faithfulness \citep{marks2024sparsefeaturecircuitsdiscovering}: 
\begin{equation}
    F = \frac{m_C-m_\emptyset}{m-m_\emptyset}.
\end{equation}
We aim to attain a faithfulness of 1: $F<1$ implies that we have missed task-relevant edges, while $F>1$ suggests we have missed edges that work \emph{against} the model's task abilities (but may nonetheless be task-relevant). In practice, circuit faithfulness trades off with size; it starts at 0, and quickly grows towards 1 as the circuit's size grows, but only reaches 1 after many less-important edges are added. As a result, it is common to study circuits with faithfulness near but below 1, which omit less important components.

\subsection{Finding Circuits}\label{sec:circuit-finding}
A naive approach to finding circuits is to compute how much the model's performance decreases when each edge is corrupted; this is the edge's indirect effect (IE; \citealp{pearl2001indirect}). Note that IE can be positive, which implies that the edge promotes model performance, or negative, implying that it harms it. To find a circuit of size $n$, one could simply take the $n$ edges with the highest $|\text{IE}|$; we take the absolute value because we want to capture all edges that affect model behavior, positively or negatively.\footnote{Past work has found components and edges that actively and systematically work against model performance \citep{wang2023interpretability}, which are a relevant part of our model's mechanism.} However, as testing an edge requires one forward pass, this approach takes $O(|E|)$ forward passes, and both $|E|$ and the cost of a forward pass increase with model size. 

We thus opt to linearly approximate each edge's IE via edge attribution patching (EAP; \citealp{syed-etal-2024-attribution}). Given an edge $(u,v)$, EAP estimates its IE as:
\begin{equation}\label{eq:eap}
    \hat{\text{IE}}_{u,v} = (z_u'-z_u)^\top\nabla_{v_{in}}m(s),
\end{equation}
where $(z_u'-z_u)$ indicates the change in the activation of $u$ upon corruption, and $\nabla_{v_{in}}m(s)$ is the change in metric when the input to $v$ changes. Computing the activations $z_u'$ and $z_u$ of all nodes $u$ requires two forward passes, while computing $\nabla_{v_{in}}m(s)$ for all nodes $v$ requires one backward pass. We can thus compute $\hat{\text{IE}}$ for all edges in a constant number of forward and backward passes, although the cost of each pass grows with model size. 

As EAP is often inaccurate in practice, we instead use \citeposs{hanna2024have} EAP with integrated gradients (EAP-IG). EAP-IG improves upon EAP by computing $\nabla_{v_{in}}m$ at intermediate points between $s$ and $s'$, interpolating between the two in representation space. Having thus computed each edge's IE, we find the circuit by taking the top-$n$ edges by $|\hat{\text{IE}}|$. We take edges by the absolute value of $\hat{\text{IE}}$ to find a circuit that is a \emph{complete} explanation of the model's task behavior, containing all relevant nodes and edges, even if they harm performance. 

Recent work in circuit-finding instead learns a mask indicating which model nodes and edges are in the circuit; the mask is trained to optimize both sparsity and model performance when the out-of-circuit units are corrupted \citep{chintam-etal-2023-identifying,bhaskar2024finding,li2024optimal}. Though we prefer techniques that provide IE estimates for each edge, mask-based approaches could be an interesting avenue for future work.

\subsection{Why use circuits to localize formal and functional linguistic competence?}\label{sec:why-circuits}
We argue for the use of circuits to localize formal and functional linguistic competence because they capture mechanisms that are both necessary \textit{and sufficient} for models to perform formal and functional linguistic tasks. Many techniques can identify parameters or neurons that harm model performance when ablated; however, such approaches may miss units that are causally relevant to the model's behavior on formal and functional tasks, as one can harm performance without ablating all relevant units.

These issues have played an important role in prior work attempting to localize a language network in LLMs. \citet{zhang-etal-2024-unveiling-linguistic}, for example, localize a ``core linguistic region'' consisting of model parameters that harm model abilities cross-lingually when set to zero. However, such zero ablations can cause harm unrelated to the importance of the units ablated (see \Cref{sec:causal-localization}); moreover, they do not check that this zero-ablation harms language selectively. Thus, we cannot be sure whether this network is crucial for LLMs' language abilities, or their abilities in general.

\citet{alkhamissi2024llm} engage in a more neuroscientifically grounded study, identifying ``language network neurons'' using a localizer task: \citeauthor{alkhamissi2024llm} select those neurons whose activations on non-word lists and on sentences differ most significantly. They find that these neurons' activations better predict brain data than random neurons do; moreover, zero-ablating these neurons harms model performance. Based on this, the authors argue that these neurons constitute a language network.

We argue that this inference, too, is flawed. As before, zero-ablation is a destructive technique, prone to harming model performance by throwing its activations out of distribution. However, this issue is exacerbated by the activation-difference localization method. It is known that certain outlier neurons have magnitudes up to 20x larger than others in the same layer \citep{timkey-van-schijndel-2021-bark,dettmers2022gpt3int8,arash2023intriguing}; such neurons are prone to be found by activation difference methods due to their high magnitudes in general. Moreover, even just quantizing these neurons (i.e. imprecisely recording their values, not zeroing them) is disastrous for model performance \citep{lin2024aqw}. These results can thus be explained by the detection and ablation of outlier neurons, rather than a language network.

In contrast to the methods of these past studies, and others that do not engage in causal analysis at all \citep{kisako2025representational}, circuits provide causal guarantees about the correctness of the desired localization. They not only use more principled ablation methods, but also aim to be necessary, sufficient, and minimal. This allows us to compare the localizations found via circuits without worrying that these include unnecessary components, or miss necessary ones.

\section{Tasks and Data}\label{sec:tasks}
In our main experiments, we consider 10 tasks that give us broad coverage over most of the subdomains of formal and functional linguistic competence described by \citeauthor{mahowald2024dissociating}. We exclude some such categories (like phonology) as they are impossible to test in text-based LMs. For others (social reasoning), LM abilities are generally poor, while circuits are best found for tasks models perform well; LM are quite competent on all tasks we study here (see App. A 
for details). See \Cref{tab:tasks} for an overview of tasks. While we introduce some tasks that are new to circuit analysis, most have previously been studied in the circuits literature. We note, however, that because prior work studied these tasks in smaller or older models, it is difficult to make direct comparisons between the circuits we find, and the circuits prior work found.

\begin{table*}[]
    \centering
    \begin{tabular}{c|p{30mm}|p{18mm}|p{60mm}}
         & \textbf{Task} & \textbf{Category} & \textbf{Input and [Expected Output]} \\
         \hline
         \hline
         \multirow{5}{*}{\rotatebox{90}{Formal}} & Subject-Verb Agreement (SVA) & Syntax & The keys on the cabinet [are] \\
         \cline{2-4}
         & Gendered Pronoun Agreement & Syntax & Maria said that [she] \\
         \cline{2-4}
         & Negative Polarity Items (NPIs) & Syntax & The customer that the managers liked has [never] \\
         \cline{2-4}
         & Hypernymy & Lexical \ \ \ \ Semantics & Roses are a type of [flower] \\
         \cline{2-4}
         & Wug Test & Morphology & First, there was one \textit{trult}. Now there are two [\textit{trults}] \\
         \hline
         \hline
         \multirow{5}{*}{\rotatebox{90}{Functional}} & Indirect Object Identification (IOI) & Situation Modeling & Alice and Bob went to the store. Then Alice gave a bottle of water to [Bob] \\
         \cline{2-4}
         & Entity Tracking & Situation Modeling & The apple is in Box F, the computer is in Box Q,\ldots, Box F contains the [apple] \\
         \cline{2-4}
         & Colored Objects & Situation Modeling & On the table, I see an orange textbook, a red puzzle, and a purple cup. What color is the textbook? \\
         \cline{2-4}
         & Greater-Than & Formal Reasoning & The war lasted from the year 1842 to the year 18[43, 44, \ldots, 99] \\
         \cline{2-4}
         & Country-Capital & World Knowledge & France, whose capital, [Paris] \\
    \end{tabular}
    \caption{Tasks under study. The top five tasks are formal, while the bottom five are functional.}
    \label{tab:tasks}
\end{table*}

\subsection{Formal Tasks}
\noindent \textbf{Subject-Verb Agreement} (SVA) is a \textit{Syntactic} task that gives models inputs like $s=$ ``The keys to the cabinet'', and expects verbs that agree with the subject \emph{keys}. Its corrupted variant inverts the plurality of the subject; here \emph{keys} would change to \emph{key}. We measure task performance as the probability assigned to verbs that agree with the subject, minus that assigned to verbs that do not. For this task, we adapted \citeposs{newman-etal-2021-refining} SVA data. SVA is a classic form of syntactic agreement, and has been used to study the language network in humans \citep{fedorenko2020selectivity}.\\

\noindent \textbf{Gendered-Pronoun Agreement} is a \textit{Syntactic} task that gives models inputs involving explicitly gendered entities, like $s=$ ``The heroine went home because''. The expected output is the corresponding gendered pronoun, \emph{she}. Its corrupted variant replaces the subject with the corresponding opposite-gender noun (here, \emph{hero}); task performance is measured as ${logit}(\text{she})-{logit}(\text{he})$. \citet{vig2020causal} identified neurons causally responsible for models' ability to perform this task in a gender-biased scenario; we adapt their data, adding explicitly gendered entities. Gendered-pronoun agreement is also a classic form of syntactic agreement, which has often been used to study the language network \citep{rodd2010functional,fedorenko2020selectivity}.\\

\noindent \textbf{Negative Polarity Item} (NPI) usage is a \textit{Syntactic} task that gives models inputs that do or do not license NPIs, like \textit{ever} or \textit{any}. For example, the input $s=$ ``The customer that the managers liked has'' could be continued by ``never'', but not by the NPI ``ever''. In each corrupted variant, we add an NPI-licensing word to $s$, or remove it if it already exists; in our example, $s'$ could be ``\emph{No} customer that the managers liked has'', which licenses the use of the NPI \emph{ever}. Task performance is measured as ${logit}(\text{never})-{logit}(\text{ever})$. We adapted this task's data from the corresponding SyntaxGym task \citep{gauthier-etal-2020-syntaxgym}. For a discussion of NPI in the context of formal semantics in the brain, see \citet{panizza2012formal}.\\

\noindent \textbf{Hypernymy} is a \textit{Lexical Semantic} task  that gives models inputs like $s=$ ``Roses are a type of'', and expects outputs like \textit{flower}. Its corrupted variant replaces the hyponym (\emph{roses}) with that of another type (e.g. \emph{diamonds}); the resulting metric is $p(\text{flower}|s)-p(\text{gem}|s)$. We use the task data from \citet{hanna2024have}. We note that this task and its categorization as a formal (lexical semantic) task, rather than a functional (perhaps world knowledge) task, may be somewhat controversial. This classification stems from \citeposs{mahowald2024dissociating} taxonomy, which specifies that lexical (and compositional) semantics are formal linguistic competences; other aspects of semantics fall under world knowledge. Hypernymy is a classic example of a lexical semantic relation, along with hyponymy, synonymy, and antonymy \citep{cruse1986lexical}; we thus consider it to be a lexical semantic and formal task. While past work in neurolinguistics has not studied hypernymy specifically, \citet{fedorenko2020selectivity} operationalized lexical semantics via one of its sister relations, synonymy. In their study, participants determined if two sentences' meanings were the same, when one word was replaced by its synonym; they observed that the language network responded during this task, providing evidence for synonymy being a formal task.\\

\noindent The \textbf{Wug Test} is a \textit{Morphological / Syntactic} task that tests models' abilities to generate singular and plural forms of nonce words \citep{berko1958child}. It gives models inputs containing nonce words like $s=$ ``First, there was one \textit{trult}. Now there are two'', and expects outputs like \textit{trults}. Its corrupted variant reverses  the number of entities present, as in $s'=$ ``First, there were two \textit{trults}. Now there is one''. Task performance is measured as $p(\text{trults}) - p(\text{trult})$. We generate new nonce words using Wuggy \citep{keuleers2010wuggy} to avoid LMs having previous exposure. The Wug test requires morphological abilities (namely, to generate the plural or singular of a nonce word) but also requires syntactic abilities (to understand that the nonce word and preceding numeral must agree in number). Past work has studied morphology's role in the language network, though it studied verbal, rather than nominal morphology \citep{bozic2010bihemispheric}.\\

\subsection{Functional Tasks}
\noindent \textbf{Indirect Object Identification} (IOI) is a \textit{Situation Modeling / World Knowledge} task providing inputs like ``When Mary and John went to the store, John gave a bottle of milk to'' and expecting the output ``Mary''. Its corrupted version replaces the second instance of \emph{John} with an unrelated name like \emph{Bob}. Task performance is measured via the difference in the logit assigned to the correct vs. incorrect entity, i.e., ${logit}(\text{Mary})-{logit}(\text{John})$. This task, common throughout the circuits literature, requires models to recognize that, if an individual has an object, they cannot give it to themselves. This can be solved via situation modeling, which entails recognizing that Mary is the other entity in this situation, along with the commonsense knowledge that people seldom give things to themselves; in \citeposs{mahowald2024dissociating} framework, this is considered world knowledge. In neuroscience, such commonsense world knowledge has been found to exist outside the language network \citep{ivanova2021language}. This task's data is adapted from \citet{wang2023interpretability}, who introduced it and studied its circuit in GPT-2 small \citep{radford2019language}.\\

\noindent \textbf{Entity Tracking} is a \textit{Situation Modeling} task that gives models inputs like $s=$ ``The apple is in Box F, the computer is in\ldots the document is in Box Q. Box F contains the'' and expects outputs like ``apple'' \citep{kim-schuster-2023-entity}. In the corrupted version, the queried object (e.g. \emph{computer in Box Q}) is different. Task performance is measured via ${logit}(\text{apple}|s)-{logit}(\text{computer}|s)$. In this task, highlighted by \citeauthor{mahowald2024dissociating} as an example of situation modeling, LLMs must track entities and their state over the length of a discourse; past work in neuroscience has studied the requisite situation modeling skills \citep{baldassano2017discovering}. \citet{prakash2024finetuning} studied this task in Llama-7B \citep{touvron2023llama}.\\

\noindent \textbf{Colored Objects} is a \textit{Situation Modeling} task that gives models inputs like $s=$ ``On the table, I see an orange textbook, a red puzzle, and a purple cup. What color is the textbook?''; the expected output is ``orange''. In the corrupted version, object colors and the queried object (e.g. ``blue mug'') are different. Task performance is measured via ${logit}(\text{orange}|s)-{logit}(\text{blue}|s)$. Much like the preceding task, the Colored Objects task requires the LLM to model the situation at hand and recall facts about it, making it a situation modeling task; past work has localized situation modeling abilities to the brain's default network, outside the language network \citep{bruckner2019brain}. \citet{merullo2024circuit} first studied this task in GPT-2 medium \citep{radford2019language}, showing that its circuit uses many of the same mechanisms as IOI does. We adapt this task's data from the original BigBench task \citep{srivastava2023beyond}.\\

\noindent \textbf{Greater-Than} is a \textit{Formal Reasoning} task with inputs like $s=$``The war lasted from the year 1842 to the year 18''; we expect outputs greater than 42. Performance is measured via $\sum_{y> \texttt{YY}}p(y|s) - \sum_{y\leq \texttt{YY}}p(y|s)$. The corrupted input for this task replaces the start year \texttt{YY} (in $s$, \texttt{YY}=``42'') with ``01''. Corrupted inputs thus shift the model's output distribution towards early years like ``02'' or ``03''; these are in general $\leq$ \texttt{YY}, as the start year \texttt{YY} in clean inputs ranges from ``02'' to ``98''. We adapt this task's data from \citet{hanna2023how}, who introduced it, and studied its circuit in GPT-2 small. Performing the comparison involved in this task requires functional competence; for similar work in humans, see \citet{amalric2019distinct}, who study math presented both in symbols and in natural language.\\

\noindent \textbf{Country-Capital} is a \textit{World Knowledge} task that gives models inputs like $s=$ ``France, whose capital,'' and expects outputs like ``Paris''. Its corrupted variant replaces \emph{France} with another country like \emph{Italy}; task performance is then measured via ${logit}(\text{Paris}|s)-{logit}(\text{Rome}|s)$. We also include the reverse task (\textbf{Capital-Country}), which provides a city, and asks which country it is the capital of; past work has shown that this has a similar circuit to the Country-Capital task \citep{hanna2024have}. Both of these tasks involve factual / world knowledge, handled by the brain's knowledge and reasoning systems, exterior to the language network \citep{Fedorenko2024TheLN}.

\section{Experimental Pipeline}\label{sec:experiments}
Our experimental pipeline works as follows. For a given model, we find the circuit for each task in \Cref{sec:tasks}, using 500 examples per task. Then, for each pair of task circuits, we measure their similarity; if a formal-functional dissociation is to exist, formal circuits should be dissimilar from functional ones but similar to one another. Here, we discuss our choice of models, circuit-finding methods, and metrics to measure circuit similarity.

\paragraph{Models} We study state-of-the-art models from five families: Llama-3 8B \citep{grattafiori2024llama3}, Gemma-2 2B \citep{gemmateam2024gemma2}, Qwen-2 7B \citep{yang2024qwen2}, Mistral-v0.3 7B \citep{jiang2023mistral7b}, and OLMo 7B \citep{groeneveld-etal-2024-olmo}. We choose these models, lying in the 2-8 billion parameter range, because they are the largest models for which current circuit-finding techniques can function, due to both memory and compute constraints. Moreover, these models are capable enough to perform some situation modeling and world knowledge tasks, on which smaller models generally fail. Note that these are base models that underwent no instruction tuning, as this type of model was analyzed by \citeauthor{mahowald2024dissociating} and is commonly studied in the circuits literature. We speculate that results for instruction-tuned models would be similar, as recent work has shown that learned features are similar across base and instruction-tuned models \citep{kissane2024saes}, and that fine-tuning mostly enhances existing circuits \citep{prakash2024finetuning}.

\paragraph{Circuit Finding}
For each task, we estimate the IE of each edge in the model's computational graph using EAP-IG, as described in \Cref{sec:circuit-finding}. Each node in the computational graph represents an attention head, an MLP, and the model's inputs or logits, and edges indicate causal links between nodes. Given these IEs, we can then construct a circuit by taking the top-$n$ edges with the highest absolute IE. For each task, we search for the minimum $n$ such that the top-$n$ circuit has a faithfulness of at least 85\%, i.e., we find the minimum $n$ such that the circuit recovers at least 85\% of the full model's task performance. In this way, we find a circuit that explains almost all of the model's performance on the task, without including a long tail of low-IE nodes and edges.\footnote{See Appendix D for results with a higher threshold, 90\%, yielding the same results as with the lower, 85\% threshold.}

Because faithfulness trades off with size, and we find circuits of a fixed faithfulness, each task's circuit may be of a different size. That is, some tasks may only rely on small circuits, while others may require the inclusion of much larger mechanisms in order to achieve 85\% faithfulness. This can make ascertaining the similarity of two circuits rather challenging, as not all similarity metrics treat different-sized circuits the same.

\paragraph{Metrics} Given two tasks $T_1, T_2$ with circuits (subgraphs) $C_1$, $C_2$, how can we compute their similarity? Past work \citep{csordas2021are} has used metrics such as intersection over union (\textbf{IoU}) and \textbf{recall} (with respect to $C_1$):
\begin{equation}
    \text{IoU}(C_1, C_2) = \frac{|C_1\cap C_2|}{|C_1\cup C_2|}, \quad\quad \quad \text{recall}(C_1, C_2) = \frac{|C_1\cap C_2|}{|C_2|}.
\end{equation}

Note that IoU and recall behave differently when circuits are of different sizes. If $|C_1| >> |C_2|$, IoU penalizes this heavily, as it is capped at $|C_2|/|C_1|$. While recall$(C_2, C_1)$ is also capped at $|C_2|/|C_1|$, recall$(C_1, C_2)$ can be as high as 1; indeed, large circuits may naturally recall more edges from smaller ones.

In contrast to these, \citet{hanna2024have} suggest measuring \textbf{cross-task faithfulness}, by running the circuit $C_1$ on task $T_2$, and vice versa. A circuit that captures many of the mechanisms required for another task should have high performance on it. In the following experiments, we use all three of these metrics; however, the appropriate metric in a given scenario depends on the hypothesis being tested.

We do note that many other potential graph-based similarity metrics exist. For example, given two task circuits, and corresponding IEs for each edge, we can compute their weighted graph edit distance between the two by summing the difference in scores assigned to each edge by each task; note that as IEs have different ranges per task, these must be normalized first. Similarly, we can concatenate the edge scores of each task into a vector, and apply metrics such as cosine similarity to these. These metrics could be useful for future work; however, pilot experiments using them yielded very similar scores for all circuits, likely due to the poor IE estimates provided by EAP and EAP-IG, which are known to capture the ordering of edges better than they capture the edges' actual IEs \citep{syed-etal-2024-attribution,hanna2024have}.

One potential concern in designing overlap-based metrics for circuit similarity is superposition, along with polysemanticity. Superposition is typically discussed as the phenomenon in which LLMs represent more features than they have dimensions (or neurons) \citep{elhage2022superposition}. Since the number of features greater than the number of neurons, some neurons must become \textit{polysemantic}: they fire on many different features in the input, rather than always firing on or representing just one feature. Later work has observed that this phenomenon is not constrained to neurons: attention heads, too, can be polysemantic \citep{kissane2024interpreting}. This raises questions, namely: if two task circuits overlapped, but this overlap were due only to polysemanticity, would our methods catch this? And does this matter? 

It is possible that two task circuits could overlap due to polysemanticity, and our methods take no explicit steps to exclude overlaps of this type. Consider, though, that the ideas underlying superposition come from neuroscience: past work has hypothesized that the human brain also represents more features than it has neurons \citep{olhausen1997sparse}. Furthermore, the localizer methodology used to find the language network also contains no safeguards against overlap due to superposition. Despite this, there is no consistent overlap between the language network and other networks across tasks and individuals. For this reason, we hold LLMs to a similar standard, requiring there to be no overlap between formal and functional circuits, irrespective of superposition.

\section{Do formal and functional networks overlap in LLMs?}

In this section, we investigate the similarity of circuits for formal and functional linguistic tasks based on the degree to which the circuits' edges overlap. In \Cref{sec:iou-main}, we conduct an analysis of circuit overlap based on intersection over union. Then, we investigate whether functional task circuits may contain formal task circuits (\Cref{sec:functional-contain-formal}). Finally, in \Cref{sec:qualitative-overlap} we characterize the edges that are shared between circuits.

\subsection{Main experiment} \label{sec:iou-main}
From a circuit overlap perspective, if formal linguistic competence forms a consistent network in LLMs, distinct from that of functional linguistic abilities, there should be no overlap between formal and functional task circuits, and high overlap between pairs of formal task circuits. We measure overlap using IoU, as it is a symmetric (non-directional) metric; here, we care about whether there is any overlap, not the direction in which the overlap occurs. 

\begin{figure*}
    \centering
    \includegraphics[width=0.59\textwidth]{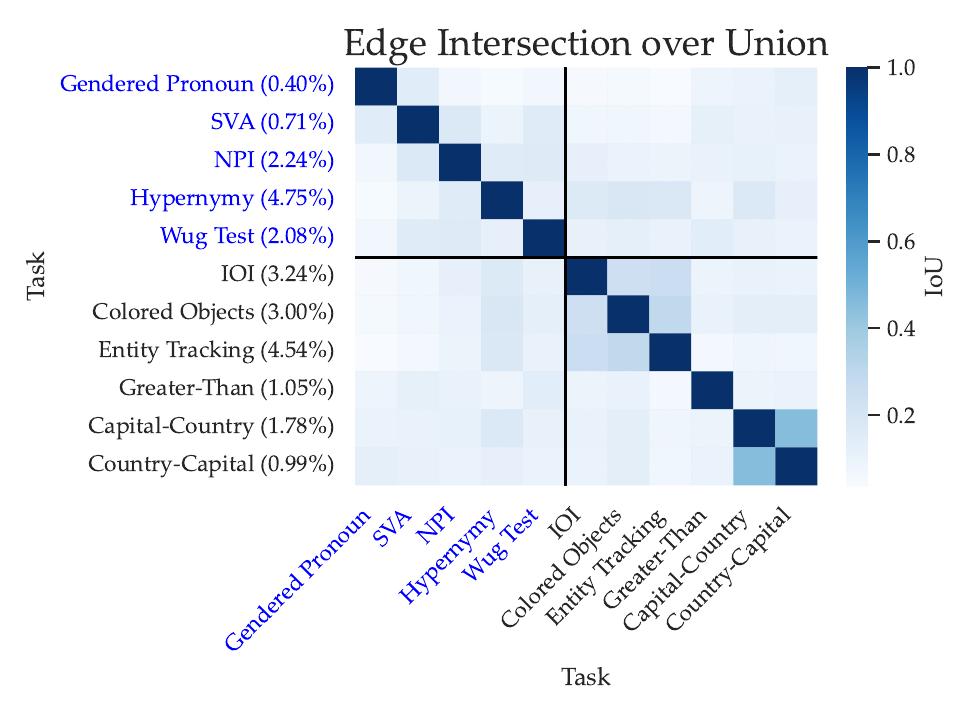}
    \includegraphics[width=0.4\textwidth]{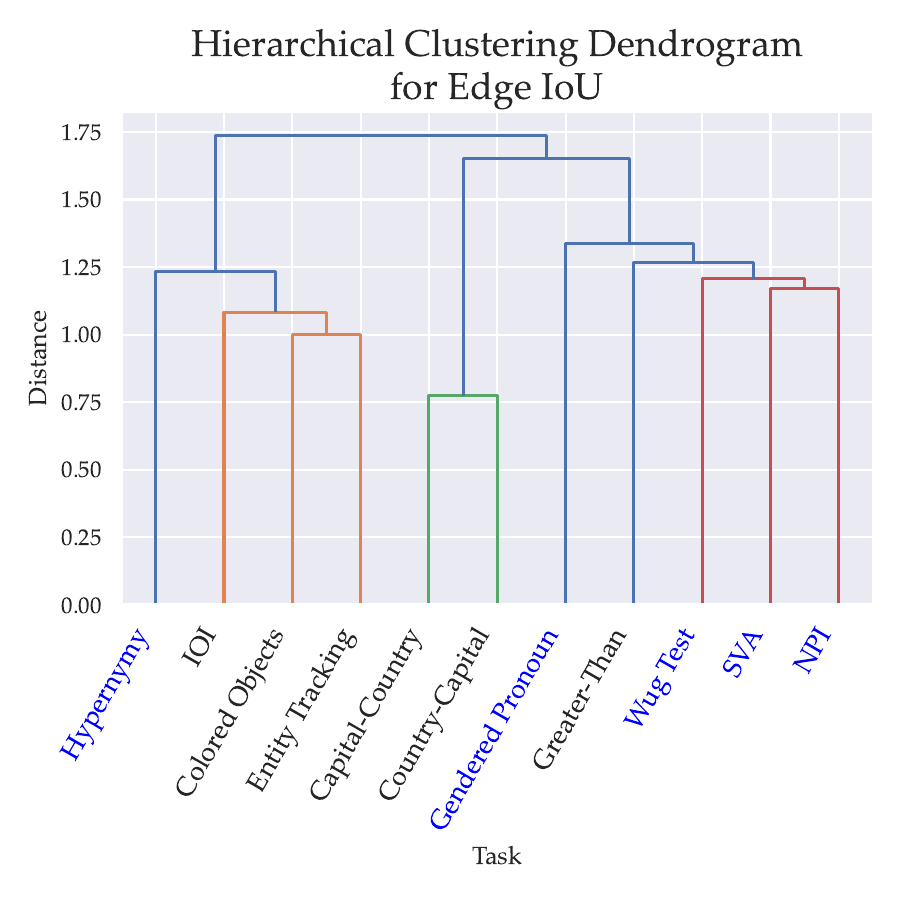}
    \caption{\textbf{Left}: Edge intersection over union (IoU) between tasks, averaged across models; as IoU is symmetric, this heatmap is symmetric as well. The average size of each task's circuit (as a percent of the entire model) is given in parentheses after each task's name. The IoU between most task pairs is low but non-zero. Moreover, formal tasks do not exhibit a higher level of overlap with one another than they do with functional tasks. Lines divide \blue{formal} and functional tasks. \textbf{Right}: Dendrogram obtained via hierarchical (agglomerative) clustering of task IoU vectors using Euclidean distance. Tasks whose IoU vectors are more similar to one another are linked at lower levels of the dendrogram. Overall, clusters do not reflect the formal-functional divide.}
    \label{fig:edge-iou-all}
\end{figure*}

In \Cref{fig:edge-iou-all}, we report the IoU between each pair of task circuits, averaged across models. We also report the average percentage of the model's edges included in each circuit, as the circuits vary in size: the SVA circuit contains only 0.71\% of model edges, on average, while the Colored Objects task contains over 4x more, at 3.00\%.  

The results indicate that in general, the IoU between any two pairs of circuits is low, but non-zero; the median IoU between tasks is 0.11. That is, while there are no striking examples of any formal and functional tasks having a high IoU, they are also not completely disjoint. Furthermore, the overlap between formal circuits is not especially high: while the median IoU between two distinct formal circuits (0.15) is higher than the median IoU between two distinct functional circuits (0.11) and between two formal and functional circuits (0.11), this difference is small.

Whether these IoUs are (statistically) significant depends on how we model a random, baseline circuit. Past work has considered random circuits with $n$ edges constructed by selecting them uniformly at random \citep{hanna2024have,shi2024hypothesis}. In this case, we can model the probability of two such circuits having an overlap of a given size using a hypergeometric distribution, and near all of the overlaps in \Cref{fig:edge-iou-all} are significantly higher than chance; see Appendix B for details on this and other techniques for modeling random overlap, all of which judge these IoUs to be significant.

Few tasks have even moderate overlap. Capital-Country and Country-Capital have high IoU (0.43), likely because they are nearly the same task; they share the same structure (fact-retrieval) and domain (geography). Similarly, IOI, Colored Objects, and Entity Tracking all have moderately high overlap (an average IoU of 0.27), perhaps because they all involve situation modeling; \citet{merullo2024circuit} also found that the IOI and Colored Objects circuits overlap. Note that even these moderate values are relatively high for edge IoU, as rather similar task circuits may only have up to 0.3--0.4 IoU; see Appendix C for more details on the ranges of edge IoU values for similar circuits.

We can verify these groupings found by visual inspection using clustering. We use agglomerative clustering to find which groups of tasks are closest to one another, as measured by the Euclidean distance of their IoU vectors (i.e. the rows or columns of \Cref{fig:edge-iou-all}, left). We then create a dendrogram using the results of this clustering. The resulting dendrogram (\Cref{fig:edge-iou-all}, right) shows that the Capital-Country / Country-Capital grouping and the IOI / Colored Objects / Entity Tracking groupings are also found by the clustering algorithm. It also finds a purely formal cluster that is less visible on the heatmap: Wug Test / SVA / NPI. However, the broader clusters do not reflect a formal-functional split. The formal Wug Test / SVA / NPI cluster next merges with Greater-Than (functional), and the functional IOI / Colored Object / Entity-Tracking cluster merges with Hypernymy (formal).

It is notable that, of all the formal tasks, hypernymy is the one that is misgrouped. This is in some sense consonant with its controversially formal status. While \citeauthor{mahowald2024dissociating} and similar researchers might categorize such lexical semantic relations, this is not a universal view. Others find less of a distinction between lexical semantics and, e.g., world-knowledge semantics \citep{hagoort2004integration}, or even view hypernymy as belonging to the latter group, more related to reasoning than lexical semantics \citep{benn2023language}. From these perspectives, the misgrouping of hypernymy could stem from the fact that it is not a formal linguistic task at all, but a functional one. However, we note that this account leaves unexplained the fact that Greater-Than, an uncontroversially functional task, is grouped with the formal tasks.

Ultimately, the results of this experiment suggest that \textbf{there is low but non-zero overlap between circuits for formal and functional linguistic abilities}. At the same time, the \textbf{formal tasks do not have especially high overlap with one another}. Taken together, these two facts weigh against a potential formal-functional distinction. 
However, there remain open questions: could the overlap between formal and function task circuits stem from the fact that both categories of tasks are expressed in language? Moreover, what is the nature of the overlap between the formal and functional tasks? If that low overlap corresponds to a task-agnostic shared processing mechanism, akin to low-level visual systems in the brain, the formal-functional hypothesis would be more plausible.

\subsection{Do functional task circuits contain formal task circuits?}\label{sec:functional-contain-formal}
Tasks involving functional linguistic competence can sometimes be presented to humans and localized in the brain without the use of language; \citet{ivanova2021language}, for example, use an image-based localizer to find brain regions responsible for event semantics. However, this is less true for LLMs: all of our functional tasks are presented via linguistic input.\footnote{Though note that this is consistent with \citet{mahowald2024dissociating}, who discuss many functional linguistic tasks presented via language to LLMs.} This could cause the functional mechanisms we localize to also include formal regions, leading to a misleading appearance of overlap.

\begin{figure*}
    \centering
    \includegraphics[width=0.59\textwidth]{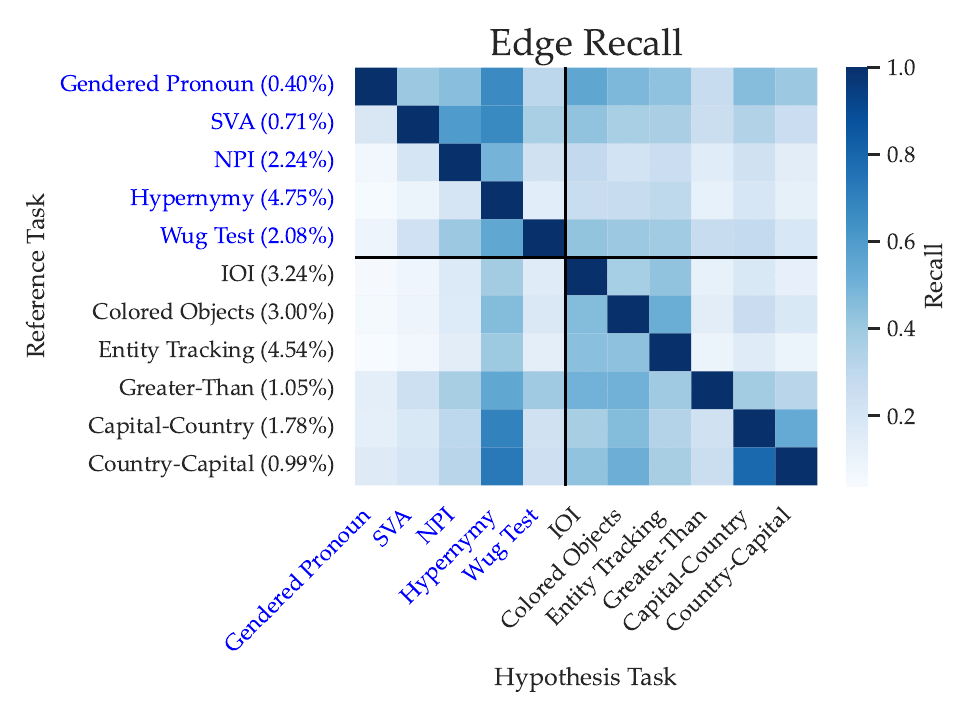}
    \includegraphics[width=0.4\textwidth]{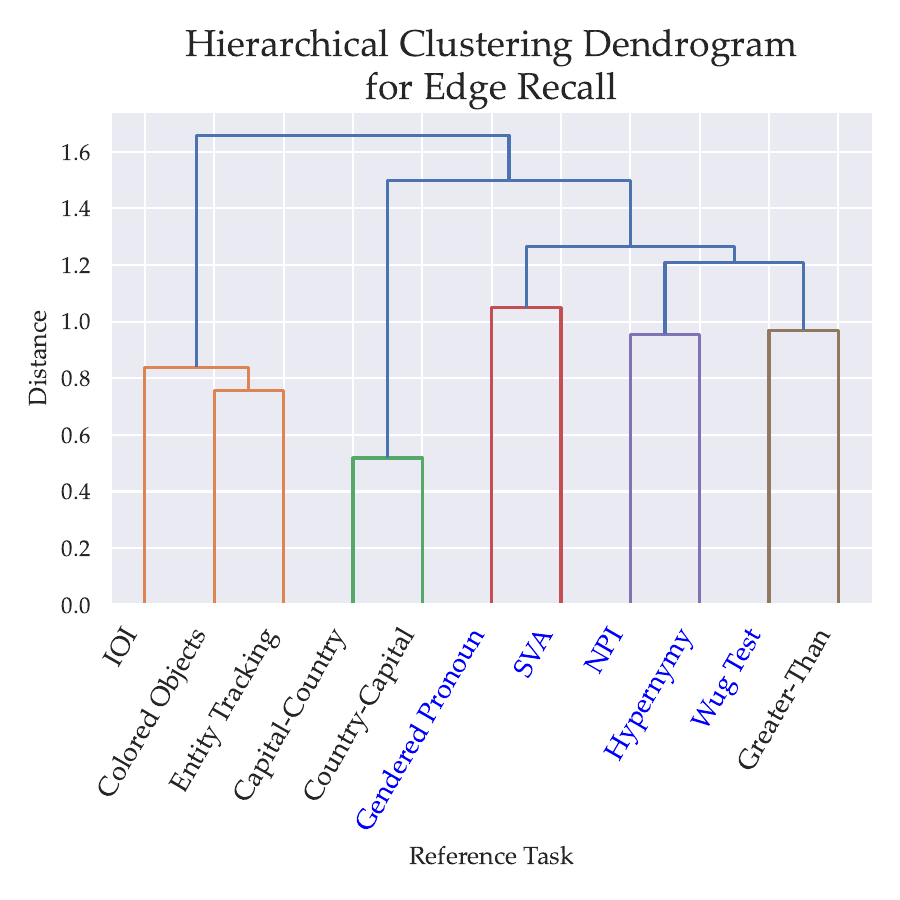}
    \caption{\textbf{Left}: Edge recall between tasks, averaged across models. Each square indicates how well the task circuit on the bottom (the hypothesis) captures the task circuit on the left (the reference). The average size of each task's circuit (as a percent of the entire model) is given in parentheses after each task's name. While functional task circuits may have higher recall of formal task circuits than formal task circuits do of functional ones, functional task circuits do not contain formal task circuits as a whole. Lines divide \blue{formal} and functional tasks. \textbf{Right}: Hierarchical clustering dendrogram for recall vectors. Formal and functional tasks cluster somewhat separately from one another, but Greater-Than is mis-clustered.}
    \label{fig:edge-recall-all}
\end{figure*}

\paragraph{Recall Analysis}
One way to demonstrate that functional tasks do not contain formal tasks is to perform the same analysis as above but measure recall, a directional measure of how much a given circuit contains another one. The results of this analysis (\Cref{fig:edge-recall-all}, left) show that functional tasks indeed do not subsume formal ones. While some formal tasks (Gendered Pronoun and SVA) with small circuits are well captured by most other task circuits, other, formal tasks like Hypernymy and NPI, are not so well captured. Moreover, Hypernymy has high coverage of many functional tasks, while the reverse is not true. In general, trends in recall seem dominated by the size of the circuits. Circuits that are large (like Hypernymy, IOI, Colored Objects, and Entity Tracking) have high recall with respect to other circuits, while being poorly covered by other circuits. Formal task circuits are not always smaller than functional task circuits: there are large formal task circuits (Hypernymy) and small functional task circuits (Country-Capital).

We also perform a clustering analysis on the reference recall vectors, using the same settings as in \Cref{sec:iou-main}; that is, the clustering should find groups of tasks that are well recalled by similar tasks. This analysis (\Cref{fig:edge-recall-all}, right) yields similarly negative results. Formal tasks cluster together, but imperfectly: the formal cluster includes Greater-Than, as well as Capital-Country and Country-Capital, more loosely.

\paragraph{Non-Language Mediated Tasks}\label{sec:non-language-mediated}

We can also test whether the small formal-functional overlap we observe is due to our posing functional tasks in language, by analyzing versions of our functional tasks that are purely functional, without any language involved. We thus create non-language-mediated versions of every one of our functional tasks; we call these our \textbf{purely-functional} tasks:

\begin{itemize}
\item \textbf{IOI (Situation Modeling)}: we alter the task to give models inputs like $s=$ ``Alice Bob Alice Bob. Carrie Dylan Dylan Carrie. Mary John John'' and expect outputs like ``Mary''. That is, the model must repeat the name that has not already been repeated, as in the original IOI task. Note that models are given in-context examples, to allow them to understand the task and format. Unlike the following purely functional tasks, this purely- functional version of IOI diverges relatively strongly from the original, which lies at the intersection of various functional capabilities such as Situation Modeling and World Knowledge; the purely-functional version does not test the latter.

\item \textbf{Entity Tracking (Situation Modeling)}: we alter the task to give models inputs like $s=$ ``\textit{apple : G, computer : F, rock : A; F : computer.} ring : A, note : B, book : P, glass : W; A :'' and expect outputs like ``ring''. That is, the task consists of a list of comma-separated ``object : letter'' pairs, followed by a semicolon and letter; models must report the object paired with the letter in the list. Models are given one in-context example (shown in italics), to help them understand the task format.

\item \textbf{Colored Objects (Situation Modeling)}: we alter the task to give models inputs like $s=$ ``orange textbook, red puzzle, purple cup; textbook:''; the expected output is ``orange''. Models are given one in-context example (not shown), to help them understand the task format. Note that in this distilled version, Colored Objects is near identical to Entity Tracking (though not IOI), reflecting the similarity of the two original tasks.\\

\item \textbf{Greater-Than (Formal Reasoning)}: we alter the task to give models inputs like $s=$``1842 < 18''; we expect outputs greater than 42. \\

\item \textbf{Country-Capital (World Knowledge)}: we alter the task to give models inputs like $s=$ ``The Netherlands : Amsterdam :: France :''; we expect outputs like ``Paris''. We include an in-context example demonstrating the task format. We also include the reverse task (\textbf{Capital-Country}), using the same analogy structure, with the relation reversed.
\end{itemize}

We study all of these tasks only in Llama-3 (8B), the most capable of our models; these abstract, in-context tasks are somewhat challenging for smaller models. We find the circuits for these five tasks, and compare them to our existing task circuits using IoU, as done previously. Here, we compare formal and purely-functional tasks, to test whether the (small) overlap we observed previously was due to using stimuli in the language format. If that was the case, that overlap should disappear in this more controlled setup. See Appendix D for comparison of the original functional tasks and their purely-functional counterparts. For more purely-functional tasks, see Appendix B.

\begin{figure*}
    \centering
    \includegraphics[width=0.48\textwidth]{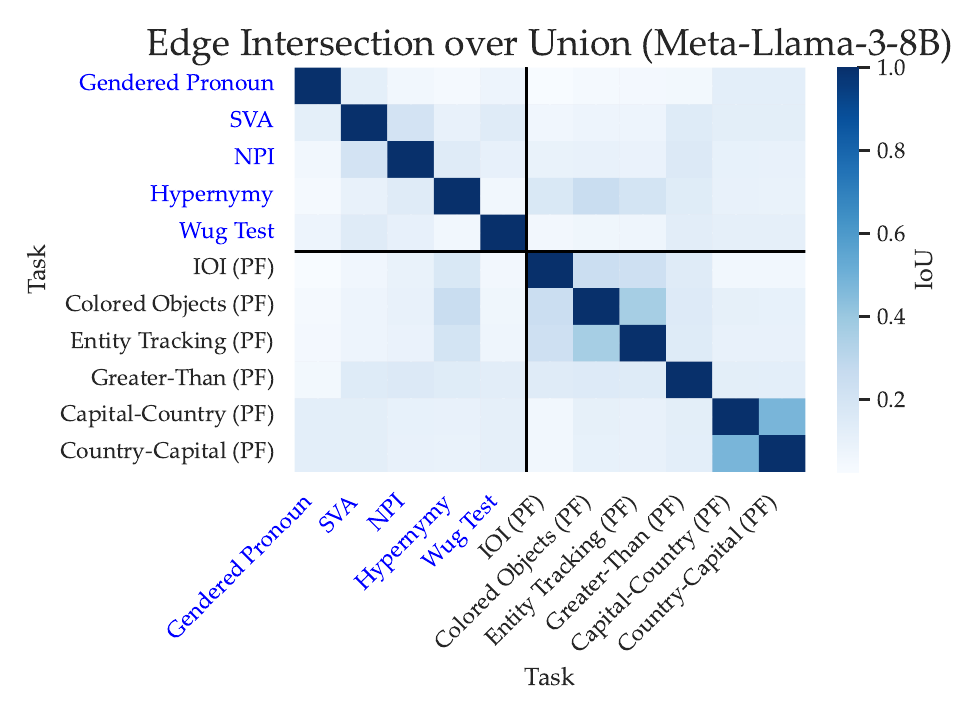}
    \includegraphics[width=0.48\textwidth]{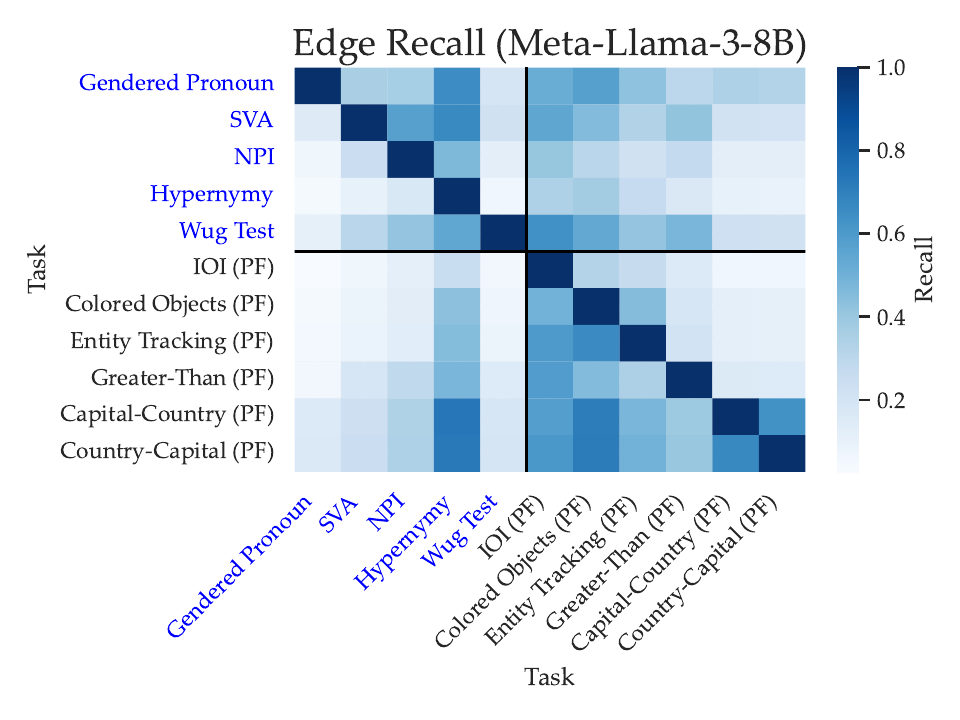}
    \caption{Edge IoU (left) and recall (right) heatmaps between formal and purely function (PF) tasks. Notably, the use of purely functional tasks has changed little, aside from weaker similarity between IOI and Colored Objects / Entity Tracking.}
    \label{fig:purely-functional}
\end{figure*}

Our results on these five purely-functional tasks (\Cref{fig:purely-functional}) are similar to those obtained using our original functional tasks. The same groupings appear: Country-Capital and Capital-Country are similar, and Colored Objects and Entity Tracking are as well, though IOI is now less similar to them. In general, IoU between purely-functional and formal tasks is still low but non-zero. We take this as evidence that the overlap between formal and functional linguistic tasks is not due to the latter containing the former.

\subsection{What components overlap between formal and functional tasks?}\label{sec:qualitative-overlap}
Our previous experiments suggest that there is formal-functional overlap, even when considering functional tasks containing no linguistic structure. But what could this overlap consist of? In order to capture entire model mechanisms for a given task, circuits must be whole paths from models' inputs to logits (\Cref{sec:circuits}). This means that, if there exist low-level task-agnostic mechanisms shared between formal and functional circuits, these might constitute a formal-functional overlap. For example, prior work suggests that transformer LMs' early layers help detokenize and contextualize words \citep{ghandeharioun2024patchscopes}. These could act as a shared low-level input processing stream, after which distinct formal or functional modules take charge.

Such low-level cross-task mechanisms also exist in the brain: brain regions responsible for vision, for example, must be active whether reading sentences or non-words. However, the localizer approach used to find the language network (\Cref{sec:language-network}) naturally avoids this issue by looking at the \textit{difference} in activations in brain regions in each condition. Reading-relevant visual brain regions should activate equally in the sentence and non-word conditions, so they are not captured by the language localizer.

We check for the existence of such a task-agnostic region by looking at the intersection of all circuits, i.e. those components and edges that are relevant across every single task that we study. We then record the types of edges in this intersection circuit---what sorts of components (the inputs, attention heads, MLPs, or the logits) do they connect, and which layers are connected? We report an average across all models.

\begin{figure*}
    \centering
    \includegraphics[width=0.49\textwidth]{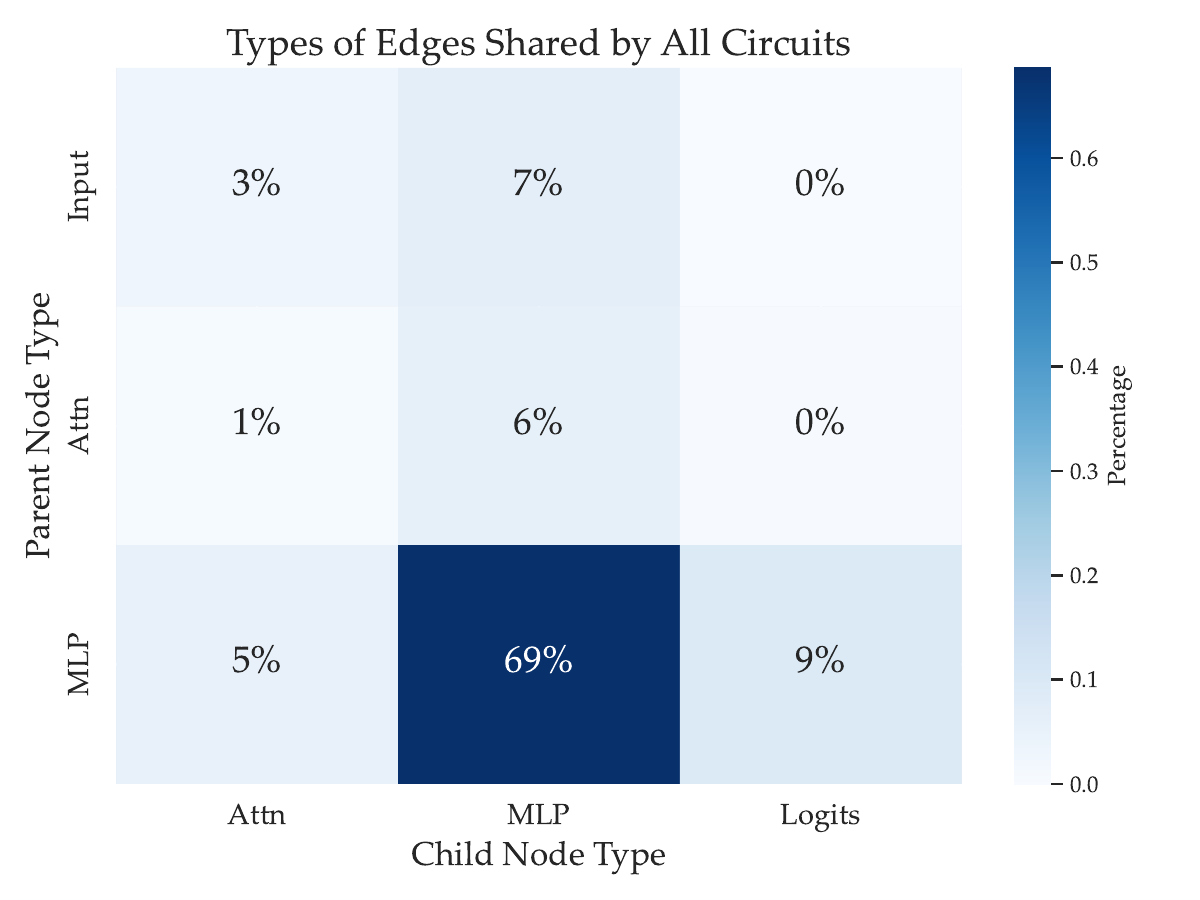}
    \includegraphics[width=0.42\textwidth]{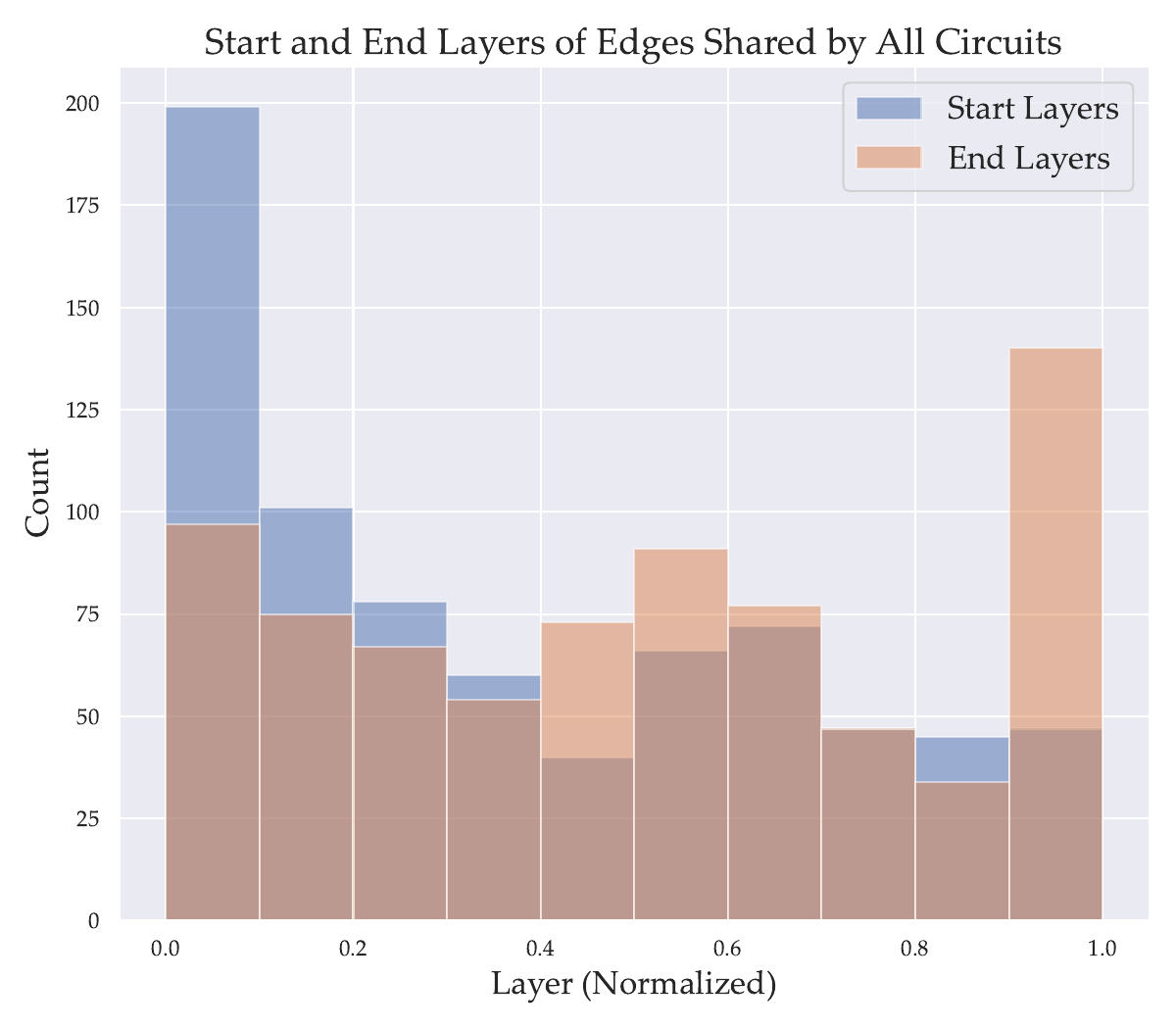}
    \caption{\textbf{Left}: Heatmap displaying the type of components (input, attention head, MLP, or logits node) connected by edges in the intersection circuit, averaged across models. Almost all edges connect one MLP to another. \textbf{Right}: Histogram of the start and end layer of edges in the intersection circuit, averaged across models. Because models have different numbers of layers, the reported layer is normalized (i.e. the actual layer of an edge's parent or child is divided by the model's total number of layers). Both start and end layers span the whole depth of the model, indicating that shared nodes and edges are not restricted to low-level (early-layer) processing mechanisms.}
    \label{fig:shared-mechanisms}
\end{figure*}

We find, much like \citet{bhaskar2024heuristic}, that there is indeed a set of the components and edges that are relevant across circuits. These are dominated by MLPs: all MLPs are included in every circuit. Our edge-level analysis (\Cref{fig:shared-mechanisms}, left) shows that the vast majority of edges shared across circuits involve MLPs. This is despite the fact that attention heads are far more numerous within all models. However, \Cref{fig:shared-mechanisms} (right) shows that the components and edges shared between circuits, including the MLPs involved, span all layers. So, insofar as a shared low-level processing area should be located at early layers of the model only (or indeed layer-wise localized at all), this hypothesis is false. Rather, this shared network seems to be composed of an ``MLP backbone'', running up and down the model, which is essential for its functioning. Unfortunately, the role of MLPs in general is rather contested; while they have been implicated in detokenization, they have also been conceived of as key-value memories for fact storage \citep{geva-etal-2021-transformer}, and implicated in various task-specific roles \citep{hanna2023how,lieberum2023does,nikankin2024arithmetic}.

Finally, when we exclude this shared network from our overlap analysis, the size of each circuit is sharply reduced: circuits' sizes fall to half of their previous size, or less. However, some overlap remains: the median IoU shrinks, but only from 0.11 to 0.10. Moreover, as we have excluded the same shared network from every circuit, the trends in IoU and recall, regarding which circuits overlap the most, remain the same. Overall, we can say that this shared MLP backbone is what most drives overlap, though it does not explain all of it.

\begin{figure*}
    \centering
    \includegraphics[width=0.59\textwidth]{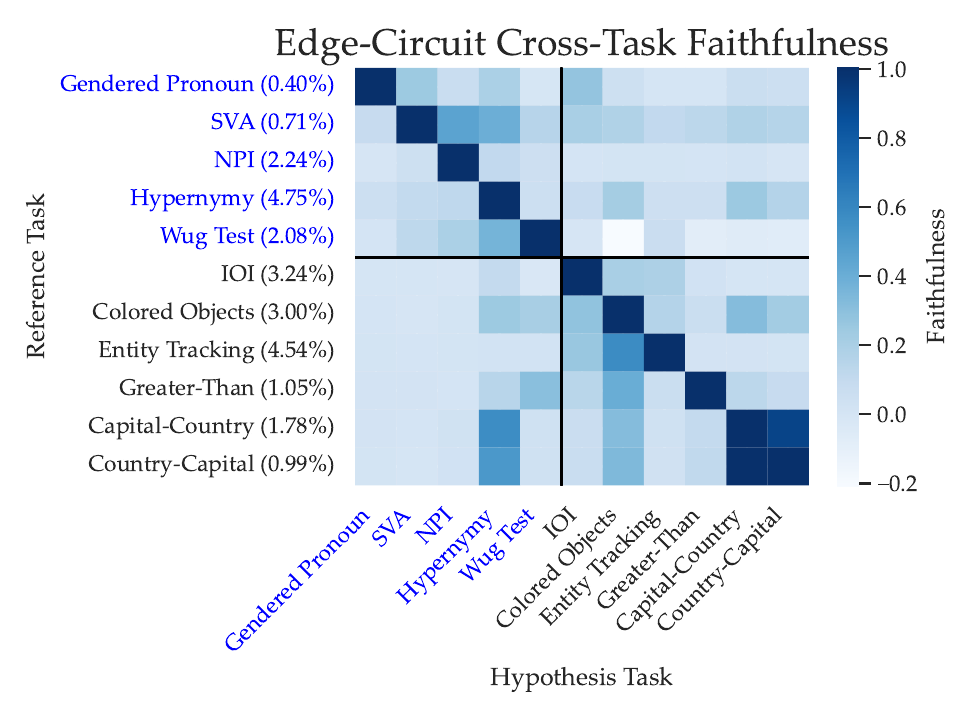}
    \includegraphics[width=0.4\textwidth]{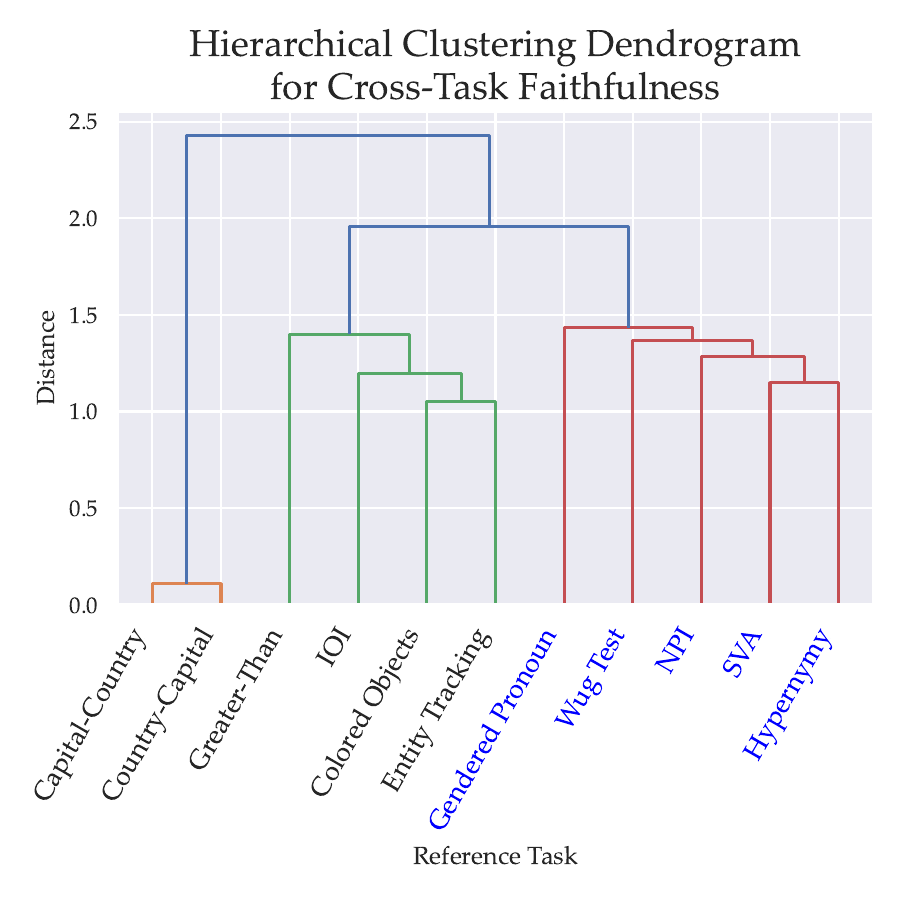}
    \caption{\textbf{Left}: Cross-task faithfulness between task pairs, averaged across models. Lines divide \blue{formal} and functional tasks. The average size of each task's circuit (as a percent of the entire model) is given in parentheses after each task's name. \textbf{Right}: Hierarchical clustering dendrogram for cross-task faithfulness vectors. Formal and functional tasks cluster separately from one another.}
    \label{fig:cross-task-faithfulness-all}
\end{figure*}

\section{Are formal and functional task circuits cross-task faithful?}
Our prior experiments provide weak evidence for a formal-functional distinction in terms of overlap. Circuits for formal and functional linguistic competence seldom overlap in today's LLMs, and what little overlap there is, is dominated by an MLP backbone. However, there does not appear to be an undifferentiated area responsible for all aspects of formal linguistic competence. In this section, we study whether these partially negative results also hold when measuring cross-task faithfulness, i.e. how well one task's circuit suffices to perform another task. If the formal-functional distinction holds, formal tasks should be able to perform other formal tasks well; moreover, they should not be able to perform functional tasks well, or vice-versa.

Examining the pair-wise cross-task faithfulness of formal and functional tasks (\Cref{fig:cross-task-faithfulness-all}, left), different trends emerge compared to earlier. While Capital-Country and Country-Capital are still similar, the IOI / Colored Objects / Entity Tracking grouping is much less clear. Moreover, while cross-task faithfulness is somewhat influenced by circuit size---the tasks whose circuits best perform other tasks (i.e., Hypernymy, Colored Objects, whose columns are dark) are large---other tasks with large circuits, like Entity Tracking do not perform other tasks as well. In general, the median cross-task faithfulness between two distinct formal tasks (0.11), or two distinct functional tasks (0.14) is higher than the median in formal-functional or functional-formal conditions (0.01 and 0.05), but this difference is still not great.

However, the results of our clustering analysis (\Cref{fig:cross-task-faithfulness-all}, right) provide more positive evidence. In this analysis, we cluster the cross-task similarity reference vectors, i.e. the vectors that show for a given task, which task's circuits solve it best. Surprisingly, all formal tasks form one cluster that is separate from all functional tasks: Gendered Pronoun, Wug Test, NPI, SVA, and Hypernymy all cluster separately from Greater-Than, IOI, Colored Objects, and Entity Tracking. We note that Capital-Country and Country-Capital form a two-task cluster that is separate from both formal and functional tasks, but this is compatible with formal-functional dissociation and formal-formal consistency. Functional tasks need not cluster together; they need only be separate from formal tasks. Overall, these results suggest that formal tasks are more alike one another with respect to which circuits solve them. So, while there is no consistent formal region at the IoU overlap level, formal task circuits are indeed more similar to each other than functional task circuits are to them at the cross-task faithfulness level. These are promising results, but to ensure they are robust, we repeat our circuit analyses at different levels of granularity in the following section.

\section{Node and Neuron-Level Circuits}\label{sec:granularity}
While our analysis centers on circuits in a computational graph composed of nodes (attention heads and MLPs) and edges, other granularities of analysis are possible. Past work has performed causal analyses of models that focus on nodes (ignoring the edges between them) or even individual neurons \citep{vig2020causal,finlayson-etal-2021-causal}. Recent work has studied \textit{sparse feature circuits} composed of features from sparse autoencoders, yet more fine-grained than neurons \citep{marks2024sparsefeaturecircuitsdiscovering}. This raises a question: might we obtain different results with a computational graph of a different granularity?

To answer this, we perform our analyses again, but at the node and neuron level. We adapt EAP-IG, which produces estimates of each edge's indirect effect, to produce estimates of each node or neuron's indirect effect. We do so by replacing $\nabla_{v_{in}}$ in \Cref{eq:eap} with $\nabla_{u_{out}}$, yielding (scalar) node IE estimates:
\begin{equation}\label{eq:eap-node}
    \hat{\text{IE}}_{u} = (z_u'-z_u)^\top\nabla_{u_{out}}m(s).
\end{equation}
If we replace the dot product above with element-wise multiplication, we obtain a vector $\hat{\text{\textbf{IE}}}_{u}\in \mathbb{R}^{d_{model}}$ of neuron IE estimates, where each entry in the vector gives the corresponding neuron's estimated IE: 
\begin{equation}\label{eq:eap-neuron}
    \hat{\text{\textbf{IE}}}_{u} = (z_u'-z_u)\odot\nabla_{u_{out}}m(s)
\end{equation}
Note that a neuron here refers to an individual dimension of token embedding, or an individual dimension of an attention head's or MLP's outputs, prior to the addition of the residual connection. This is in contrast to, e.g., a hidden neuron within an MLP.

We then compute our metrics (IoU, recall, and cross-task faithfulness) with respect to nodes or neurons, instead of edges. We omit circuits with edges between neurons due to their computational infeasibility, and sparse feature circuits because they would require sparse autoencoders for each model we study.

\begin{figure*}
    \centering
    \includegraphics[width=0.49\textwidth]{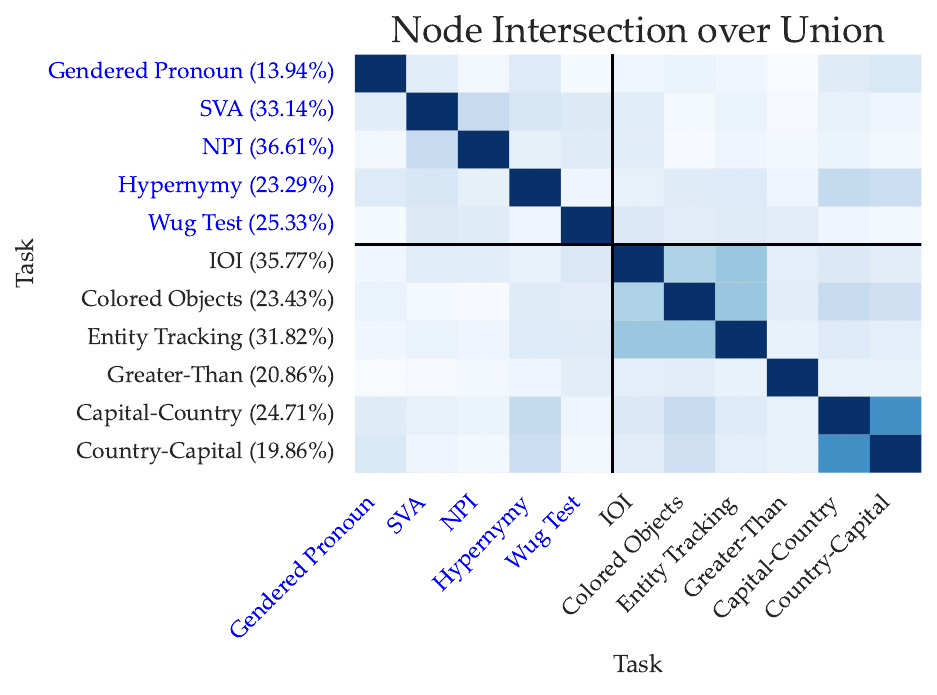}
    \includegraphics[width=0.49\textwidth]{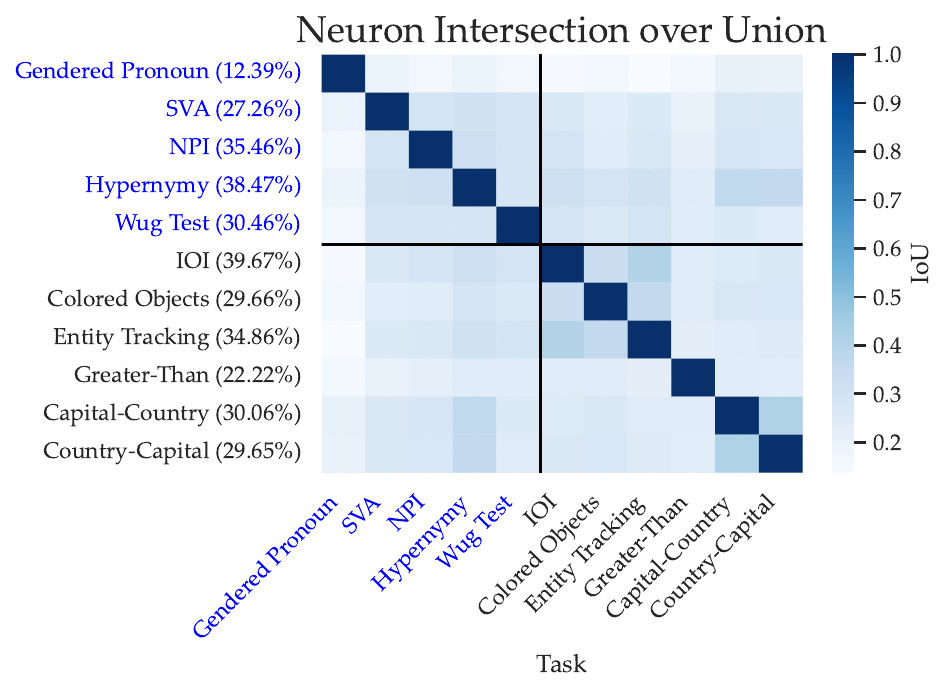}
    \caption{\textbf{Left}: Node-level IoU between task pairs, averaged across models. The average size of each task's circuit (as a percent of the entire model) is given in parentheses after each task's name. As with our edge-level experiments, most task pairs have low but non-zero IoU, and formal tasks exhibit no higher level of overlap with one another than they do with functional tasks. Lines divide \blue{formal} and functional tasks. \textbf{Right}: Neuron-level IoU between task pairs, averaged across models. IoUs are generally quite low between all tasks.}
    \label{fig:node-neuron-iou}
\end{figure*}

Our node-level IoU results (\Cref{fig:node-neuron-iou}, left) are rather similar to our edge-level results. The same circuits overlap with one another: Capital-Country and Country-Capital have highly similar circuits, while IOI, Colored Objects, and Entity Tracking are all similar to one another. All other circuits have low IoU. Overall, the circuits are much larger than in the edge scenario, including around 25\% of the model on average. This is unsurprising, as nodes are much larger units than edges; including one node is essentially equivalent to including all of its outgoing edges. While we could previously only include one edge out of a node, if that edge was important, we now must include the entire node.

Neuron circuits (\Cref{fig:node-neuron-iou}, right) contrast more with edge circuits. The increased granularity of the computational graph in the neuron case does not enable greater sparsity; neuron circuits are more similar in size (in terms of percentage of the whole model) to node than edge circuits. While some of the same trends in task overlap are visible as in past IoU analyses, neuron circuits have low IoU in general. Thus, neither IoU analysis provides further evidence for the existence of a formal linguistic region.

\begin{figure*}
    \centering
    \includegraphics[width=0.49\textwidth]{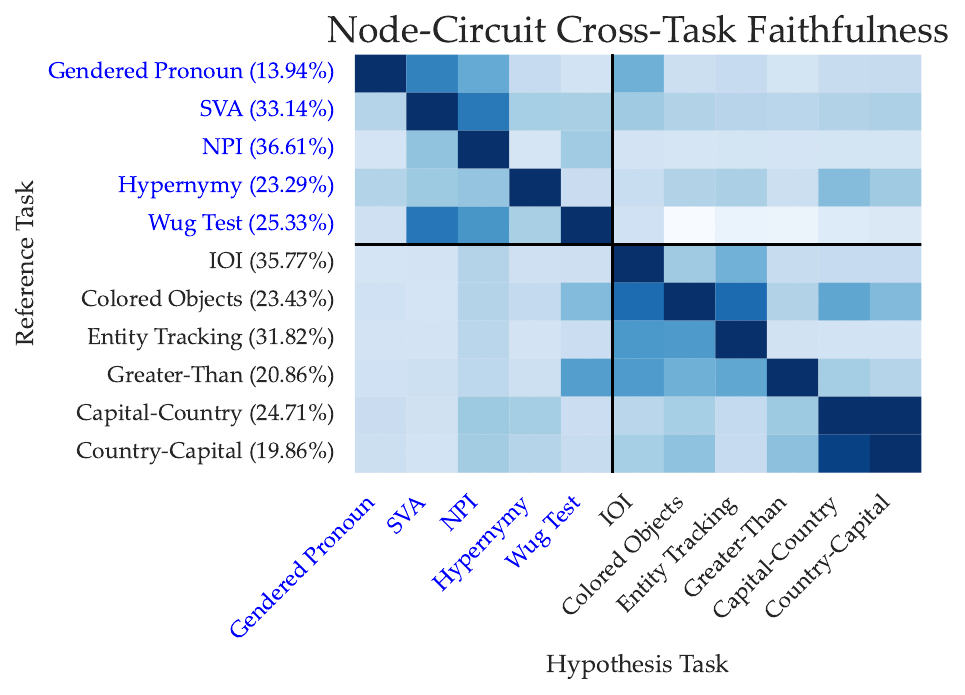}
    \includegraphics[width=0.49\textwidth]{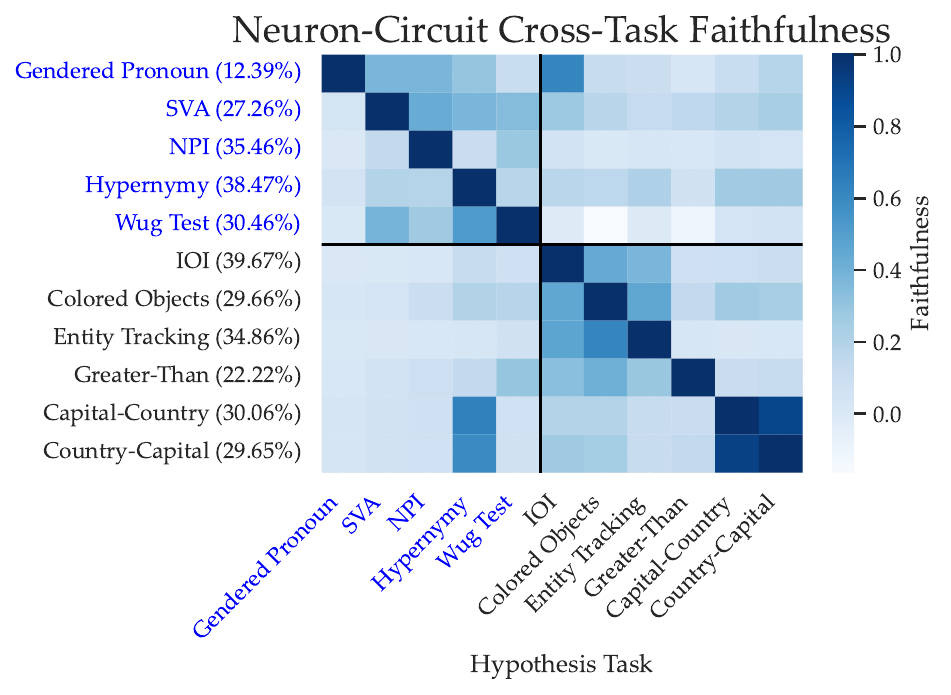}
    \caption{Cross-task faithfulness for node (\textbf{left}) and neuron (\textbf{right}) circuits. The average size of each task's circuit (as a percent of the entire model) is given in parentheses after each task's name. In both cases, formal tasks clearly have higher cross-task faithfulness with one another than with functional tasks, and vice-versa. Lines divide \blue{formal} and functional tasks.}
    \label{fig:node-neuron-cross-task}
\end{figure*}

However, the cross-task faithfulness of node and neuron circuits (\Cref{fig:node-neuron-cross-task}) are more suggestive of a formal-functional distinction. In particular, formal tasks have a higher cross-task faithfulness on other formal tasks than on functional tasks, while functional tasks capture one another much better than they capture formal tasks. On the node level, the NPI and SVA circuits capture not only each other, but also the Wug Test and Gendered Pronoun tasks relatively well; on the neuron level, formal-formal similarity is more generalized, but still visible. 

\begin{figure*}
    \centering
    \includegraphics[width=0.49\textwidth]{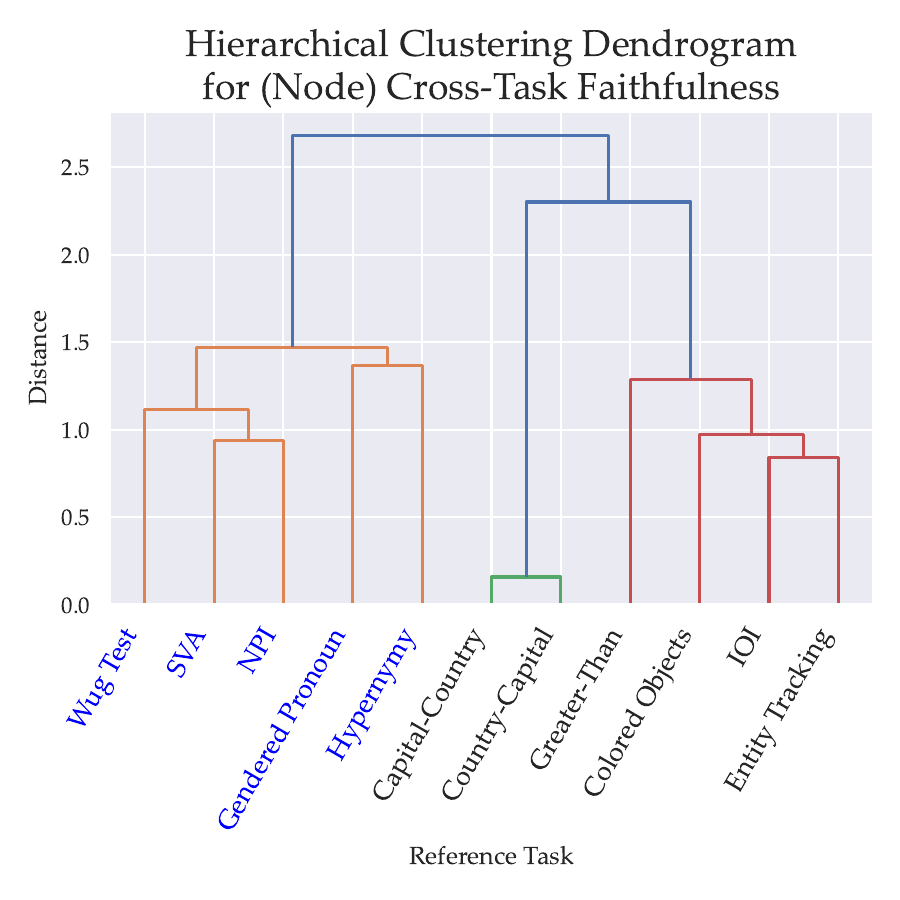}
    \includegraphics[width=0.49\textwidth]{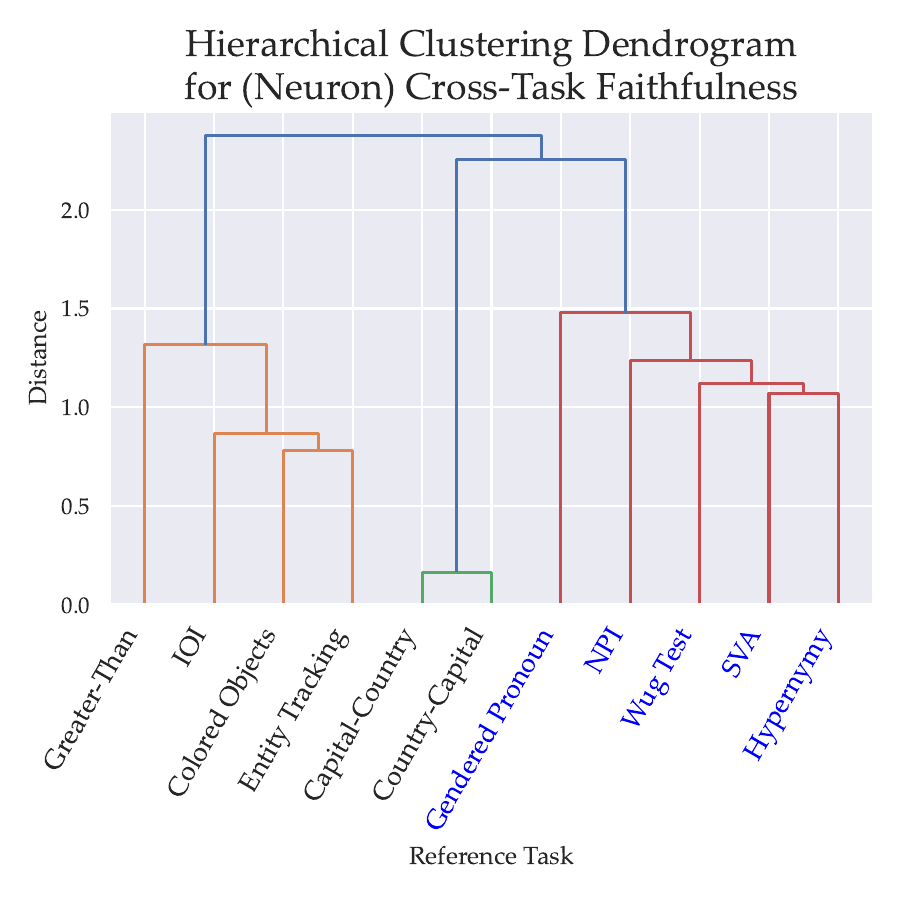}
    \caption{Cross-task faithfulness dendrograms for node (\textbf{left}) and neuron (\textbf{right}) circuits. In both cases, formal tasks clearly cluster together. \blue{Formal} tasks are labeled in \blue{blue}, functional tasks in black.}
    \label{fig:node-neuron-cross-task-dendrogram}
\end{figure*}

A clustering analysis at either level (\Cref{fig:node-neuron-cross-task-dendrogram}) yields the same conclusion. At both the node and neuron level, there is a clustering of the formal tasks as separate from the functional ones. The consistent emergence of a formal grouping as measured by cross-task faithfulness provides stronger evidence for a division between formal and functional task mechanisms in LLMs. However, the formal grouping that we uncover is defined not by the precise edges, nodes, or neurons that fall within it, but by the general ability of formal tasks to perform other formal tasks. 

\section{Discussion}\label{sec:discussion}
In this paper, we have translated \citeposs{mahowald2024dissociating} hypothesis about emergent formal-functional dissociation in LLMs into the circuit analysis framework, and tested the hypothesis. We have tested this in two ways, first measuring if formal and functional circuits contain overlapping units, and then measuring the cross-task faithfulness between the two. Our results indicate that formal and functional circuits have low but non-zero overlap. Moreover, formal circuits also have low overlap in general with one another, suggesting that there is no undifferentiated formal language region in LLMs as in the human brain. Formal circuits are also not a subset of functional ones in general, even when these functional tasks are posed using language. These results are stable across different circuit granularities and faithfulness thresholds. However, when we measure cross-task faithfulness, a different pattern emerges. At the edge level, we see some clustering of formal tasks as separate from functional tasks. Then, at the node and neuron level, a stronger clustering of such tasks emerges, with formal tasks clustering entirely separately from functional ones. 

Which of these, overlap or cross-task faithfulness, should we trust? Overlap most closely parallels the methods used in the fMRI studies supporting the idea of the language network in the brain: first, one localizes the regions of the brain with a given function and then checks if they are the same. However, we argue that the appropriate measure for mechanistic similarity in LLMs may differ somewhat from this. 

The ideal metric for mechanistic similarity should measure not only \textit{if the same units are involved in a given pair of tasks}, but also \textit{if they are involved to the same extent}. This could be achieved by using, e.g. a weighted recall metric, where one task's recall of another's edges is weighted by the IE that the latter assigns to each edge. Even better metrics could compute quantities such as the graph edit distance between the entire IE-scored computational graphs for two different tasks; this sort of metric would remove the need for choosing a circuit of a fixed size, and instead directly compare the causal relevance of every unit between the two tasks. However, as discussed in \Cref{sec:experiments}, such metrics are hampered by the fact that EAP and EAP-IG estimate IEs relatively poorly in absolute terms \citep{syed-etal-2024-attribution}.

How, then, can we take into account the causal importance of each unit in our model when measuring mechanistic similarity? Cross-task faithfulness measures precisely this: the more a given circuit captures causally important units for a given task, the better that circuit will perform on the task. Moreover, capturing units with a higher IE should naturally yield a larger increase in cross-task faithfulness than capturing units that are causally relevant, but have a low IE. So, we view cross-task faithfulness as a valid and important way to measure mechanistic similarity as well. Indeed, cross-task faithfulness has parallels in humans. \textit{Dual-task interference} studies ask subjects to perform two tasks at the same time; if their performance suffers, one infers that the two tasks share the same neural resources \citep{watanabe2014neural,leone2017cognitive}.

Still, even if formal-functional groupings emerge when measuring cross-task faithfulness, should this then be considered evidence for separate formal and functional modules? To answer this, we can again return to the human neuroscience literature, in which a similar phenomenon has been observed. The human brain, too, can exhibit \textit{degeneracy}, whereby multiple neural circuits lead to the same downstream behavior \citep{price2002degeneracy,mason2010degeneracy,mason2015hidden}; this is also discussed in philosophy as \textit{multiple realizability} \citep{bickle2020multiple}. Crucially, this degeneracy can be taken as evidence against modularity; if a neural module is an area of the brain that is necessary and sufficient to complete a task, then the existence of multiple circuits that are sufficient (but not on their own necessary) to complete a task provides evidence against modularity \citep{zerilli2019neural}. 

Past LLM circuit work has already shown evidence for degeneracy via the phenomenon of self-repair, where model components that initially appeared unimportant to a task take on new roles and compensate for the task's circuit, when it is ablated \citep{wang2023interpretability,mcgrath2023hydra}. If we interpret cross-task faithfulness as degeneracy, this serves as evidence against the existence of separate modules for each formal task. However, there does seem to be a collection of model components that support formal linguistic competence. Not all of these components are strictly necessary for each task; sometimes a task can be solved via multiple combinations of such components. This runs contrary to the idea of a formal linguistic network that is undifferentiated and wholely necessary for all formal linguistic competences. At the same time, we found that circuits for formal linguistic tasks are better at solving other formal linguistic tasks than functional linguistic circuits are.

Finally, it is worth noting that, throughout this study, we have operated under the assumption that the formal-functional distinction in the brain as discussed by \citeauthor{mahowald2024dissociating} is clear-cut. However, this is not uncontroversial: both the specific aspects of language that are included in the formal language network, as well as the formal-functional distinction more broadly, are both topics of heated debate. The distinction between lexical / combinatorial semantics and world knowledge is one site of conflict, where what constitutes each group, and if these groups should be separate at all, is contested \citep{hagoort2004integration, pylkkanen2009semantics}. Others reject the formal-functional distinction wholesale due to its reliance on fMRI studies, as opposed to more temporally sensitive techniques \citep{murphy2024LN}, and advocate for an approach to language in the brain that stresses the connectivity between core language regions and other cognitive systems \citep{forkel2024redefining}. While the critiques of hypernymy's categorization find some support in our edge IoU analyses, arguments against the formal-functional distinction more broadly are less supported by our analyses, which show some dissociation between the two.

Beyond our results on formal-functional dissociation, this present study also presents a much larger-scale work than most circuit analyses. In particular, recent circuit papers often study one or two tasks in GPT-2 small \citep{li2024optimal} or single models with larger parameter counts \citep{prakash2024finetuning,bhaskar2024finding}. In contrast, we study 5 different models of 2-8 billion parameters, across 10 tasks. This allows us to both build on existing results---we find that the IOI and Colored Objects tasks are not only similar to each other, as found by \citet{merullo2024circuit}, but also similar to Entity Tracking. Other results, such as the existence of an MLP backbone across circuits, and the fact that circuits have low overlap with one another in general, were possible only because of the larger models and more numerous tasks that we studied.

In conclusion, we have investigated the question of formal-functional dissociation in LLMs via a wide-ranging circuit similarity study across five formal and five functional tasks, and three granularities. In doing so we provide the first application of circuits to questions from neuroscience and discover new phenomena, such as an MLP backbone running down all circuits. However, as in prior work, we find major differences in similarity depending on whether we measure overlap and cross-task faithfulness \citep{hanna2024have}. Measuring overlap shows low formal-functional similarity, but also low formal-formal similarity; in contrast, cross-task faithfulness suggests clear formal-functional dissimilarity and formal-formal similarity. Overall, we interpret this as showing the existence of a loose collection of formal mechanisms, where circuits drawn from this set are more able to solve formal tasks than circuits drawn from outside it. Not all of these mechanisms are strictly necessary for all formal tasks; a single formal task may have multiple solutions within this flexible collection. Future work could study more tasks, in order to precisely define the limits of LLMs' formal mechanisms, and further the use of mechanistic interpretability to understand the relationship between LLMs and the brain. Future work could also study the emergence of potentially shared mechanisms during LLM pretraining.

\section{Limitations}\label{sec:limitations}
In this study, we analyzed five formal and five functional tasks, limited by the difficulty of crafting tasks that fit into the circuits framework; studying more tasks would help confirm that formal and functional tasks do indeed have distinct patterns in cross-task faithfulness. Moreover, we have studied circuits for these tasks found at two faithfulness thresholds, 85\% and 90\%; varying these thresholds further could still cause our results to change.

We note that cross-task faithfulness has another limitation: while we generally take high cross-task faithfulness to entail high similarity between two mechanisms, cross-task faithfulness rewards circuits that find components / edges that positively contribute to target task ability. However, insofar as we want to find \textit{complete} circuits (i.e., circuits that contain all causally relevant model units, even negatively acting ones), faithfulness is not always informative; high faithfulness may result from the exclusion of negatively acting units. In the circuit-finding scenario, where we try to find a 85\% faithful circuit, we attempt to avoid this by including units based on their absolute IE; in the cross-task faithfulness scenario, however, we cannot be as sure that we have found all important components. We argue that for our purposes, this is acceptable; even if formal task circuits capture primarily positive-contribution components and edges of other formal tasks, this constitutes a formal network.

\section*{Acknowledgments}
The authors would like to thank Kyle Mahowald and Anna Ivanova for helpful discussions about this paper, and Raquel Fern\'{a}ndez for reading it and providing insightful feedback. The authors also thank the Dialogue Modeling Group at the University of Amsterdam, and the Technion NLP group for their many useful comments and suggestions. We thank anonymous reviewers for directing us to the neuroscientific literature on degeneracy and NPIs.

MH is supported in part by an OpenAI Superalignment Grant. This work was conducted in part during a research visit funded by the ELIAS Mobility Program (Grant Agreement No. 101120237). This research was supported by an Azrieli Foundation Early Career Faculty Fellowship and  by Open Philanthropy. This research was partly funded by the European Union (ERC, Control-LM, 101165402). Views and opinions expressed are however those of the author(s) only and do not necessarily reflect those of the European Union or the European Research Council Executive Agency. Neither the European Union nor the granting authority can be held responsible for them.

\newpage
\appendix 
\appendixsection{Base Model Performance on Tasks}
\label{app:behavioral-performance}
In this paper, we find circuits for models' task abilities. However, we must verify that models can perform these tasks; otherwise, there may be no circuit to be found. Here, we measure baseline model performance. We measure top-1 accuracy for tasks where there is one clear correct answer: Hypernymy, Wug Test, IOI, Colored Objects, Entity Tracking, Capital-Country, Country-Capital, and all other variants studied. 

However, for some tasks (Gendered Pronoun, SVA, NPI, and Greater-Than), there is no one correct answer, and that answer need not always be the top output. Consider the case of SVA with the input \textit{The keys on the cabinet\ldots}; while \emph{are} is clearly more correct than \emph{is}, or any other singular-conjugated verbs, much of the model's probability may be distributed to words like \textit{that}. Such words are neither right nor wrong, so we should not penalize models for outputting them. In such cases, we score each example as correct if models assign more probability to the right token (or set thereof) than to the wrong token (or set thereof); alternatively, we could use accuracy with respect to the first token that is either correct or incorrect, rather than neither. Results for the tasks shared across all models are in \Cref{tab:accuracy}. Accuracies tend high, except for IOI and Entity Tracking.

\begin{table}[b]
\begin{tabular}{lrrrrr}
Task | Model & Llama 3 & Qwen 2.5 & OLMo & Mistral-v0.3 & Gemma 2 \\
\hline
Gendered Pronoun & 0.99 & 0.99 & 0.99 & 1.00 & 0.98 \\
SVA & 0.92 & 0.98 & 0.95 & 0.96 & 0.96 \\
NPI & 1.00 & 0.98 & 0.98 & 0.98 & 0.99 \\
Hypernymy & 0.98 & 0.97 & 0.97 & 0.99 & 0.98 \\
Wug Test & 0.89 & 0.87 & 0.86 & 0.93 & 0.90 \\
IOI & 0.55 & 0.72 & 0.69 & 0.65 & 0.79 \\
Colored Objects & 0.99 & 1.00 & 0.55 & 0.93 & 0.88 \\
Entity Tracking & 0.80 & 0.99 & 0.42 & 0.77 & 0.91 \\
Greater-Than & 1.00 & 1.00 & 0.87 & 1.00 & 1.00 \\
Capital-Country & 0.87 & 0.84 & 0.84 & 0.86 & 0.85 \\
Country-Capital & 0.97 & 0.95 & 0.95 & 0.96 & 0.93 \\
\end{tabular}\caption{Model accuracies on the tasks for which we find circuits. Though some tasks are more difficult than others---IOI and Entity tracking are particularly tough---accuracies generally surpass 90\%.}\label{tab:accuracy}
\end{table}

\appendixsection{Random-Chance Circuit Overlap}\label{app:baseline-overlap}
\paragraph{Modeling Overlap with a Hypergeometric Distribution}
One way to model the overlap that might happen between two circuits simply by chance is to consider a circuit whose edges are selected uniformly at random. Imagine two circuits $C_1 = (V_1, E_1)$ and $C_2 = (V_2, E_2)$, constructed in such a fashion; denote by $G=(V,E)$ the whole model's computational graph. We can model the probability that they overlap to a given degree using a hypergeometric distribution. If the intersection of the two circuits has size $k = |E_1\cap E_2|$, the probability of an overlap of exactly that size is

\begin{equation}
    p(k) = \frac{{|E_2| \choose k}{|E|-|E_2| \choose |E_1|-k}}{{|E| \choose |E_1|}}.
\end{equation}

In words, imagine that we draw $|E_1|$ edges without replacement from our total population of $|E|$ edges; $|E_2|$ edges in this population belong to $C_2$. This formula computes the probability that we draw precisely $k=|E_1\cap E_2|$ edges belonging to $C_2$. This same formula can be applied to node overlap, replacing $E, E_1, E_2$ with $V, V_1, V_2$. To compute the probability of an overlap of at least size $k$, we use the cumulative density function of the hypergeometric distribution, computed with SciPy \citep{virtanen2020scipy}.

\paragraph{Estimating Overlap with Dummy Circuits} The hypergeometric modeling approach is mathematically interesting, but a poor model of circuits. Circuits do not consist of randomly chosen edges: rather, they consist of paths from the inputs of the graph to the logits. Moreover, each node in a circuit tends to have a limited number of important, high contribution edges; that is, the distribution of edge importance is not uniform. How can we simulate the random selection of such circuits?

We do so using the following procedure. For each task, with a corresponding circuit, we construct a dummy circuit. This consists of a computation graph where each edge has a score drawn from a log normal distribution; we find that this simulates well the tendency of nodes to have just a few contributing edges. We then take the top-$n$ edges from the dummy circuit by (absolute) score, where $n$ is the number of edges used when applying top-$n$ search to find the corresponding task circuit. We prune both the task and dummy circuit, ensuring that neither has any nodes that lack outgoing or incoming edges, and that both sides of each edge are in the circuit.

We construct 20 dummy circuits for each task and model, and compute the average IoU between the task and dummy circuit. We find that the task and dummy circuits have rather low IoUs: most fall below 0.02, less than a fifth of our median IoU of 0.11 between pairs of real circuits. So, even with a more realistic model of random circuits, the overlap we find lies notably above the mean.

\appendixsection{More Purely-Functional Tasks}
We also study two additional tasks that are purely functional. The first is \citeposs{nikankin2024arithmetic} basic \textbf{Math} task, which consists of simple arithmetic problems involving addition, subtraction, and multiplication, such as \textit{25 + 3 =}. Corrupted examples share the same operator, but different operands. We measure performance via the difference in logit assigned to the clean and corrupted equations' answers. Presented in purely symbolic form, this task should engage purely functional (formal reasoning) areas of the model, if such exist. This task is highly sensitive to tokenization,\footnote{LLMs differ widely in how they tokenize numbers; while Llama-3 tokenizes numbers up to 3 digits in length as one token, other models (e.g. Gemma-2) tokenize them digit-by-digit, and others tokenize them in groups of 2. Single-token prediction tasks are most compatible with circuit analysis, so we use Llama-3.} and cannot easily be ported to new models, so we study it in Llama-3 (8B) alone.

The second is a string manipulation task called \textbf{Echo}, inspired by a task of the same name from \citet{hupkes2021compositionality}. In this task, models see 4 distinct tokens randomly sampled from the model's tokenizer, and must repeat the last token; a successfully solved example looks like ``$t_1$ $t_2$ $t_3$ $t_4$ : $t_4$''. We give the model two solved examples (2-shot), and then have it finish an incomplete example. In corrupted examples, the final token ($t_4$) is replaced with another, distinct token. We measure performance via the logit assigned to the clean and corrupted examples' final token. Since the tokens are randomly sampled, there should be no trace of linguistic structure; this task is most akin to a pattern-recognition formal reasoning task.

\begin{figure*}
    \centering
    \includegraphics[width=0.8\textwidth]{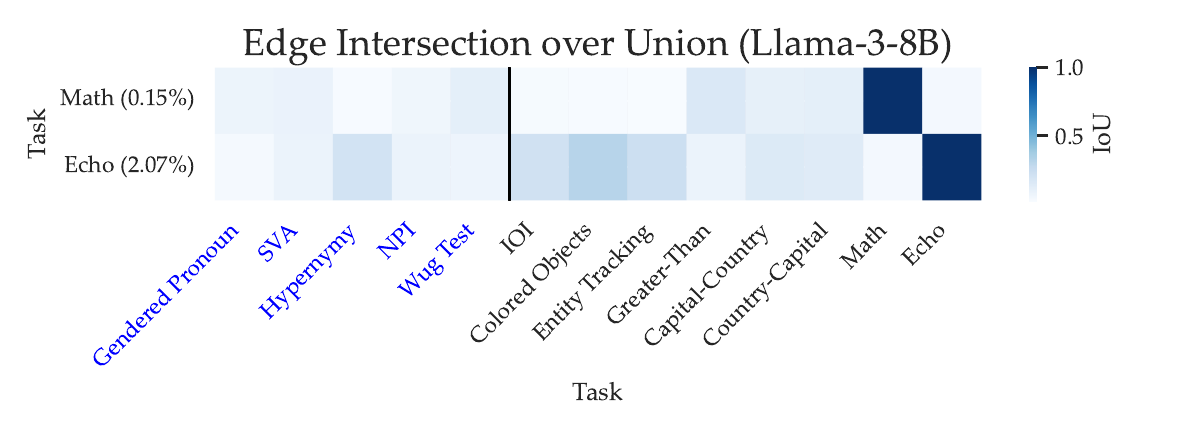}
    \caption{IoU heatmap for the Math and Echo tasks. Neither task has especially high overlap with any other task, although Echo has somewhat higher IoU with Colored Objects and Hypernymy.}
    \label{fig:math-echo}
\end{figure*}

We find the circuits for these two tasks, and compare them to our existing task circuits using IoU, as done previously. Our results on these two tasks (\Cref{fig:math-echo}) are similar to those of other functional tasks. Neither Math nor Echo is particularly similar to any other task. Notably, although one might hypothesize that they are supported by the same mechanisms, Math and Greater-Than do not have especially high overlap. Echo also overlaps slightly more with Colored Objects, IOI, Entity Tracking, and Hypernymy than with other tasks. In general, IoU between Math / Echo and formal tasks is low but non-zero (0.09 on average). We take this as evidence that the overlap between formal and functional linguistic tasks is not due to the latter containing the former.

\appendixsection{Circuit Overlap Between Similar Tasks}\label{app:similar-overlap}
How much do circuits for similar tasks overlap? In the main text, we study Capital-Country and Country-Capital, finding a high IoU (0.5) and near-perfect cross-task faithfulness. However, though these tasks seem , they need have maximally similar metric; past work suggests that models trained on ``A is B'' relations do not necessarily learn ``B is A'', indicating that these two tasks could rely on different mechanisms \citep{berglund2024reversalcursellmstrained}. Could more similar tasks have more similar circuits? And what is the range of IoU and cross-task faithfulness on non-identical circuits? In the following two sections, we compute the similarity of even more similar tasks. In general, we find that very high IoUs ($\geq0.5$) are uncommon, while very high cross-task faithfulness values are not. This suggests that even relatively low levels of edge IoU may indicate a higher degree of similarity than one might intuitively expect.

\paragraph{Greater-Than Variants} In past work, \citet{hanna2023how} found high cross-task faithfulness between the Greater-Than task and variants that phrased the task as a statement about prices (\textbf{Greater-Than-Price}: ``The price of the purse ranges from \$1842 to \$18\ldots'') or as a sequence of increasing numbers (\textbf{Greater-Than-Sequence}: ``1734, 1745, 1788, 1801, 1842, 18\ldots''). We construct the same tasks for Llama-3 (8B), and compute the edge IoU between these tasks. Surprisingly, the edge IoU ranged from 0.25 - 0.31, lower than expected for such similar tasks with high reported cross-task faithfulness.

\paragraph{Functional vs. Purely-Functional Tasks} In \Cref{sec:non-language-mediated}, we studied purely-functional versions of functional tasks, which contain no natural language structure. We can also study the overlap between similar tasks by comparing the circuits for functional tasks with those for their purely-functional counterparts. We do this, and compute (edge) IoU and cross-task faithfulness.

\begin{figure*}
    \centering
    \includegraphics[width=0.48\textwidth]{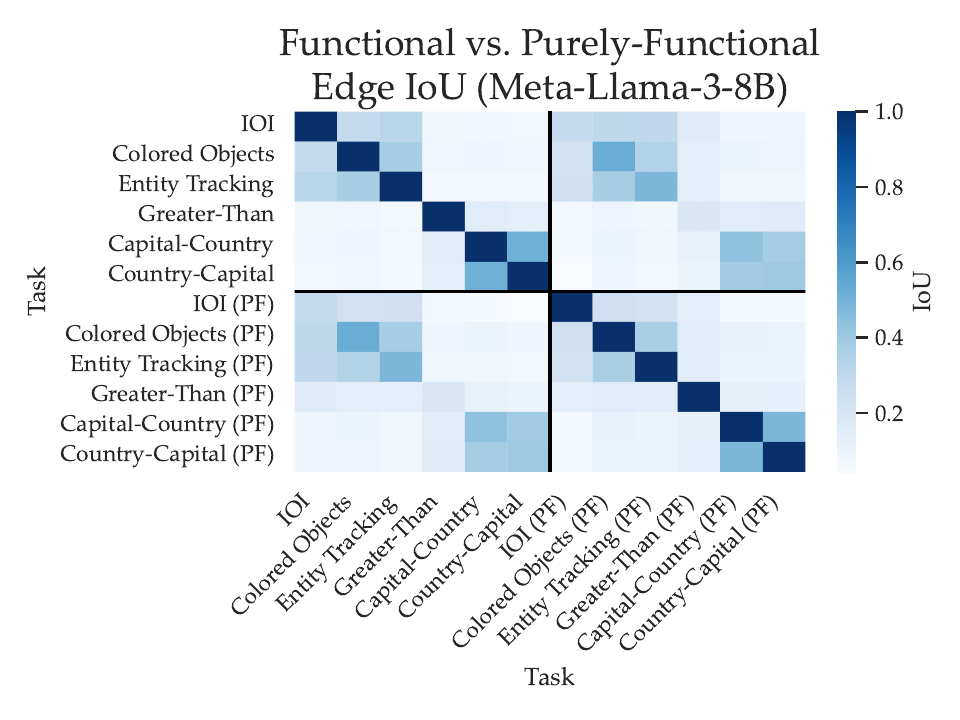}
    \includegraphics[width=0.48\textwidth]{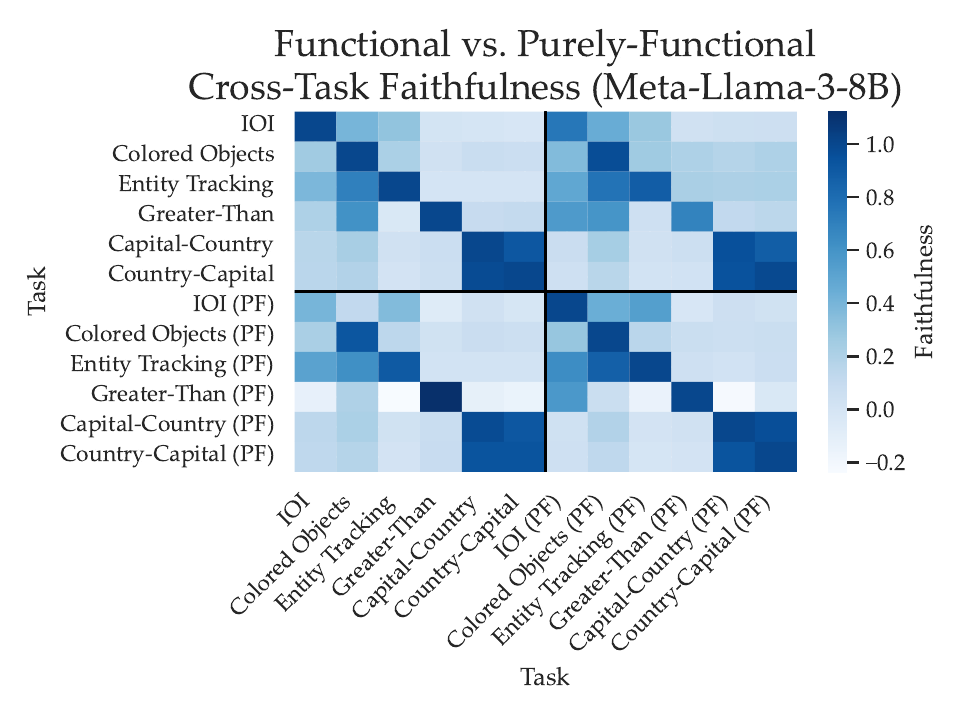}
    \caption{Edge IoU (left) and cross-task faithfulness (right) between pairs of functional tasks and their purely-functional (PF) counterparts. In all cases, there is notable similarity between each functional and purely-functional pair. However, the similarity between normal functional tasks and their purely-functional counterparts is much stronger when measured using cross-task faithfulness.}
    \label{fig:funcfunc-overlap}
\end{figure*}

Our results (\Cref{fig:funcfunc-overlap}) suggest that there is indeed overlap between functional tasks and their purely-functional counterparts. In the case of Edge IoU, this IoU is modest but clearly present: IOI, Colored Objects, and Entity Tracking overlap with their purely-functional counterparts. The same is true for Country-Capital and Capital-Country. However, the maximum IoU between a functional and purely-functional task is around 0.5. In the case of cross-task faithfulness, these trends emerge even more strongly: functional tasks have cross-task faithfulness of near 1.0 on their purely-functional counterparts. The major exception to this is IOI (PF) and the original IOI; notably, the former is a much simplified version of the latter. There are discrepancies between the two metrics: IoU indicates very low similarity between Greater-Than and its purely-functional counterpart, while cross-task faithfulness suggests that their circuits are highly similar. This highlights the differences in the two metrics. The overall difference in ranges also suggests that we should consider even small levels of IoU to be significant, given the compressed range of values it takes on.

\appendixsection{Replication with Higher Faithfulness Threshold} \label{app:higher-threshold}

\begin{figure*}
    \centering
    \includegraphics[width=0.48\textwidth]{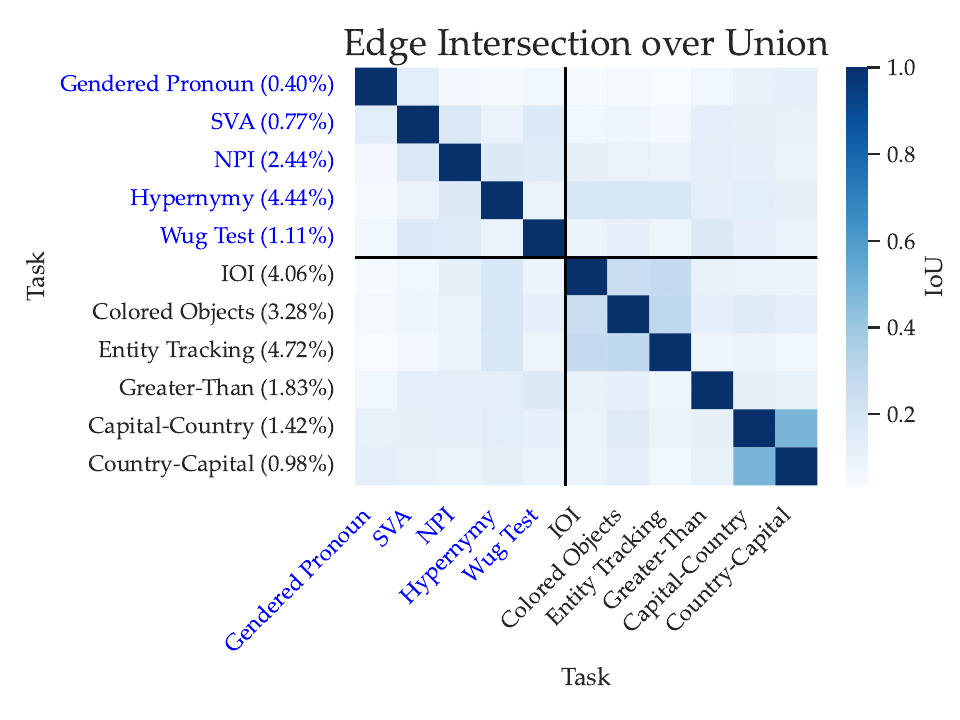}
    \includegraphics[width=0.48\textwidth]{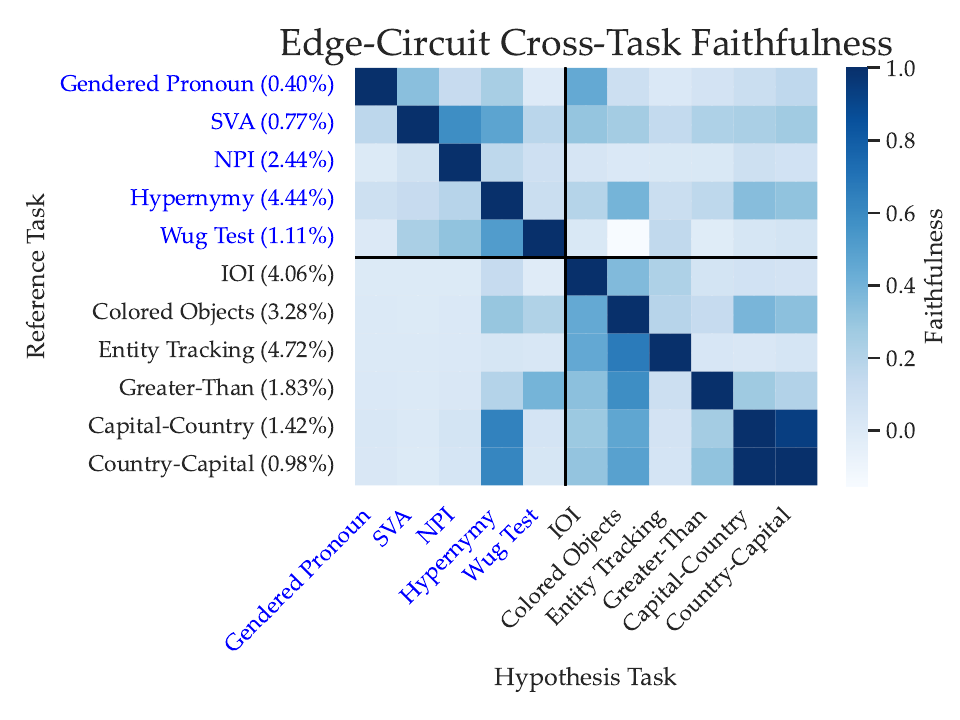}
    \caption{IoU and cross-task faithfulness heatmaps using a 90\% threshold. The same trends emerge as in the original 85\% analysis. There is low formal-functional similarity in the edge IoU case, but also low formal-formal similarity. In the cross-task faithfulness case, there is a stronger formal-formal overlap.}
    \label{fig:090-replication-heatmaps}
\end{figure*}

\begin{figure*}
    \centering
    \includegraphics[width=0.48\textwidth]{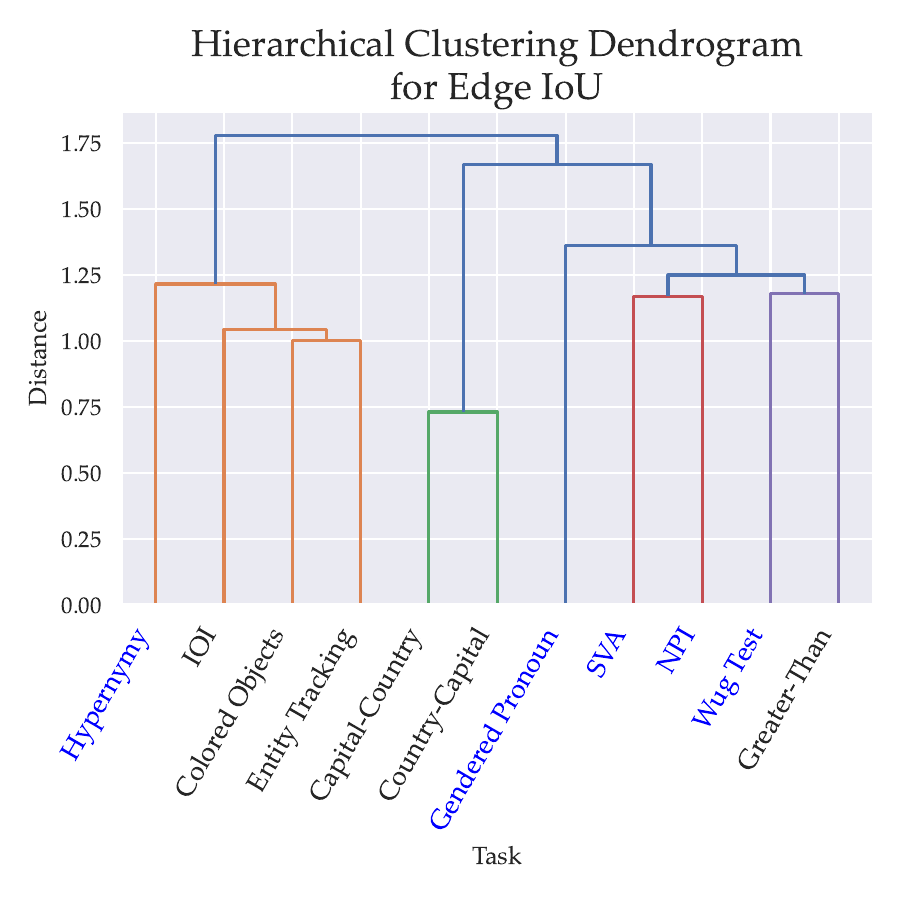}
    \includegraphics[width=0.48\textwidth]{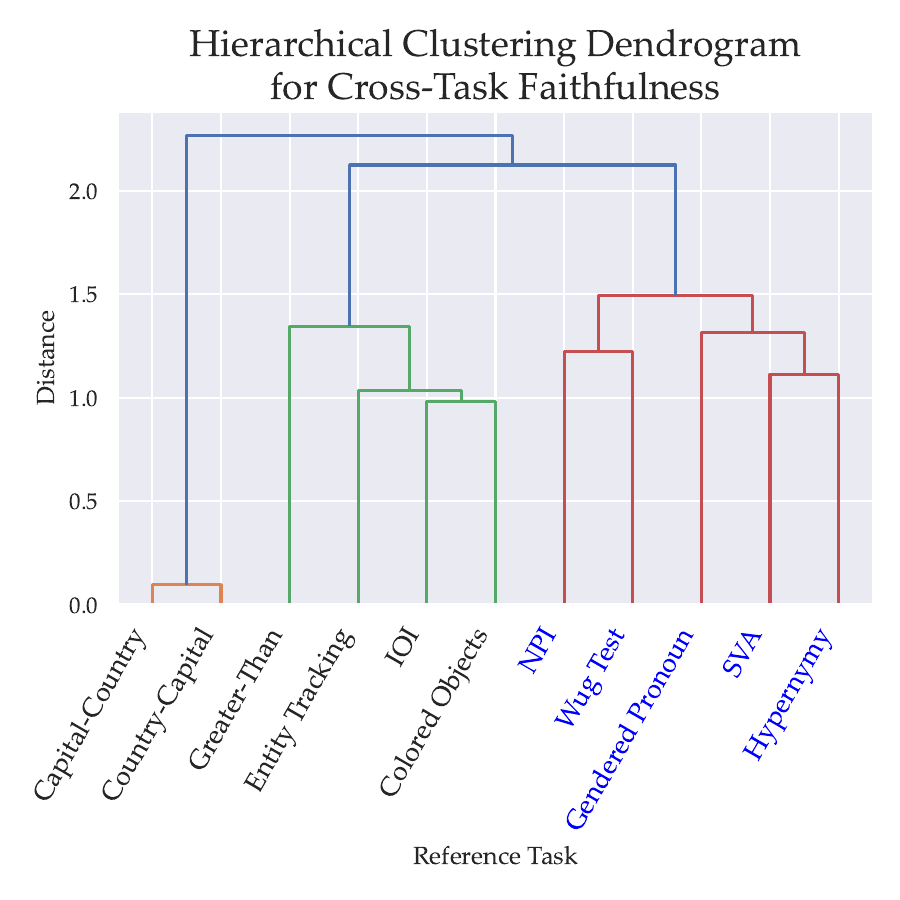}
    \caption{IoU and cross-task faithfulness dendrograms using a 90\% threshold. The same trends emerge as in the original 85\% analysis. The edge IoU dendrogram yields imperfect groupings (with Hypernymy and Greater-Than swapped), while the cross-task faithfulness dendrogram recovers the formal-functional distinction perfectly.}
    \label{fig:090-replication-dendrograms}
\end{figure*}

In order to ensure that our results are result to the chosen 85\% faithfulness threshold, we ran our experiments again, with a higher, 90\% faithfulness threshold. Notably, the results are the same, in both the heatmap (\Cref{fig:090-replication-heatmaps}) and dendrogram (\Cref{fig:090-replication-dendrograms}) analyses at the edge level, as well as at the node / neuron level (not shown).

\newpage
\newpage
\starttwocolumn
\bibliography{anthology-filtered,custom}

\end{document}